%% file: main.tex
\input{glyphtounicode}
\pdfglyphtounicode{ff}{0066 0066}
\pdfglyphtounicode{fi}{0066 0069}
\pdfglyphtounicode{fl}{0066 006C}
\pdfglyphtounicode{ffi}{0066 0066 0069}
\pdfglyphtounicode{ffl}{0066 0066 006C}
\pdfglyphtounicode{endash}{2013}
\pdfglyphtounicode{emdash}{2014}

\documentclass[10pt,twocolumn,letterpaper,fleqn]{article}

\usepackage[pagenumbers]{cvpr}

\input{preamble}

\definecolor{linkblue}{rgb}{0.21,0.49,0.74}
\usepackage[breaklinks,colorlinks,allcolors=linkblue]{hyperref}
\hypersetup{pdftitle={FIDAL: Diversity-Aware Federated Active Learning Under Real-World Distribution Shifts},
            pdfauthor={David Due\~nas Gaviria, Shadi Albarqouni},
            pdfkeywords={open-set, federated learning, active learning, out-of-distribution, medical imaging, annotation efficiency}}

\title{FIDAL: Diversity-Aware Federated Active Learning\\Under Real-World Distribution Shifts}

\author{David Due\~nas Gaviria\qquad Shadi Albarqouni\thanks{Corresponding author.}\\
University of Bonn, University Hospital Bonn,\\Clinic for Diagnostic and Interventional Radiology, Venusberg-Campus 1, 53127 Bonn, Germany\\
{\tt\small \{David.Duenas-Gaviria, Shadi.Albarqouni\}@ukbonn.de}
}

\begin{document}
\setcounter{topnumber}{5}
\setcounter{dbltopnumber}{5}
\renewcommand{\topfraction}{0.95}
\renewcommand{\dbltopfraction}{0.95}
\renewcommand{\textfraction}{0.05}
\renewcommand{\floatpagefraction}{0.8}
\renewcommand{\dblfloatpagefraction}{0.8}
\maketitle

\begin{abstract}
Federated learning enables collaborative model training across institutions without centralizing data, yet high annotation costs, domain shifts, and class imbalance remain major obstacles, especially when irrelevant out-of-distribution (OOD) samples dilute the labeled data. Existing active learning methods target uncertainty or diversity within in-distribution (ID) data and overlook unknown samples in federated clinical settings. We propose FIDAL, an open-set federated active learning framework that combines calibrated global--local evidential uncertainty, support-set diversity weighting, and adaptive OOD rejection. The rejection gate thresholds a foundation-model Gaussian-coverage signal per client and per round with Otsu's criterion, so that highly informative ID samples are queried while irrelevant outliers are excluded without any hand-tuned threshold. Evaluated on three multi-center medical imaging benchmarks (dermatology, histopathology, and mammography with organically occurring artifacts) in realistic open-set scenarios, FIDAL outperforms detector-based open-set methods by up to about 12 percentage points of balanced accuracy and is the only method on the accuracy--ID purity Pareto front of all three benchmarks. At an equal query budget it spends at least 1.3 times fewer annotations on OOD samples than every accuracy-matched baseline, saving an estimated 7--29 hours of expert reading on the mammography benchmark. By labeling only a fraction of the data pool, it matches or exceeds fully supervised performance across modalities. These results highlight the value of integrating uncertainty, diversity, and OOD rejection in open-set federated active learning for medicine.
\end{abstract}

\section{Introduction}
\label{sec:introduction}
Deep neural networks excel in medical image analysis but rely on large annotated datasets~\citep{wang2024comprehensive}, which are costly to obtain because expert annotation is slow and data are scarce. Active learning (AL) reduces labeling cost by prioritizing informative samples~\citep{settles2009active}, but its benefits diminish in federated settings, where data are distributed across privacy-sensitive institutions~\citep{pmlr-v54-mcmahan17a}. Real-world deployment adds a further complication: heterogeneous client pools inevitably contain out-of-distribution (OOD) samples arising from protocol variations, rare pathologies, or cross-site domain shifts. Such samples degrade performance when overlooked~\citep{kothawade2021similar,Kim2023DeepAL}, yet most AL~\citep{chen2024think,gal2017deep,sener2018active,ashdeep,luo2013latent,roy2001toward,li2025fast} and federated AL (FAL) methods~\citep{10378423,Chen2024FedEvi,Wu2022FederatedAL,goetz2019active,ahn2024federated,10205184} assume a purely in-distribution (ID) or generically non-identically distributed (non-IID) pool in a \textit{closed-set} setting. They therefore overlook two issues: the dominance of noisy, unknown classes from the \textit{open-set} label space, and the absence of critically informative edge cases.

Early AL relied on classical uncertainty metrics such as entropy~\citep{gal2017deep,ashdeep} and on diversity or representativeness criteria~\citep{you2014diverse,sener2018active,nguyen2004active}; transferred directly to the federated setting, both suffer from miscalibration under data heterogeneity~\citep{ahn2024federated,li2025fast}. Recent FAL approaches use Dirichlet-based evidential models to capture aleatoric and epistemic uncertainty from both the clients and the global model~\citep{chen2024think,Chen2024FedEvi}, or combine local and global models to select diverse samples~\citep{10378423,10205184}. Although these methods address class imbalance, domain shift, and non-IID data, they all assume an unrealistically pure ID pool~\citep{yu2023turning}. Open-set AL (OAL) has emerged to avoid wasting annotation effort on OOD data, first through auxiliary OOD detectors~\citep{Kim2023DeepAL,safaei2024eoal}, submodular functions~\citep{kothawade2021similar}, or contrastive models~\citep{du2021contrastive}, and more recently through detector-free single-score formulations~\citep{zong2024bidirectional,yan2024contrastive} or pre-trained vision-language priors~\citep{heo2024clipnal,zhong2025openpath}. PAL~\citep{yang2023not} further argues that selectively querying some OOD instances can sharpen the ID decision boundary. However, \emph{these methods are all centralized}, and their assumptions do not survive the move to a federated deployment (Section~\ref{sec:related}).

In this work we address \emph{Open-Set Federated Active Learning (OS-FAL)} for medical imaging, where clients face non-IID data, dynamically emerging OOD samples, and strict annotation budgets. Realistic clinical acquisition demands three properties at once: \emph{(i)} federated training across heterogeneous, privacy-sensitive clients, \emph{(ii)} open-set rejection of OOD samples before annotation, and \emph{(iii)} active acquisition under a tight labeling budget. To our knowledge, FIDAL is the first framework to address all three axes jointly within a single acquisition strategy. Our contributions are as follows.
\begin{itemize}
    \item We propose FIDAL, a unified OS-FAL framework that combines calibrated evidential uncertainty, diversity-aware selection, and OOD rejection under federated non-IID data. It uses an asymmetric \emph{global--local evidential calibration}: global epistemic uncertainty captures domain shift, while global and local aleatoric uncertainty avoid noisy samples.
    \item We introduce \textbf{support-set diversity weighting}, which goes beyond geometric diversity and explicitly promotes under-represented regions of the feature space, improving coverage of rare classes and edge cases across heterogeneous clients.
    \item We develop an \textbf{adaptive, client-specific OOD rejection} mechanism built on a foundation-model (FM) Gaussian-coverage signal $G(\mathbf{x})$ with an \emph{Otsu gate} that thresholds the unlabeled-pool likelihood distribution per client and per round, without any hand-tuned percentile. Computed in frozen FM space (PanDerm, CHIEF, or MammoCLIP), the signal acts as a task-agnostic manifold prior that is robust to real-world artifacts, and a complementary similarity penalty $\lambda_{\text{OOD}}\,S_{\mathrm{OOD}}(\mathbf{x})$ over the client's growing labeled-OOD set handles artifacts that cluster tightly enough to pass the gate.
    \item We conduct an extensive study on three federated medical benchmarks, \textbf{FedISIC} (dermatology), \textbf{FedCamelyon} (histopathology), and \textbf{FedEMBED} (mammography with organically occurring artifacts), on which FIDAL consistently achieves higher or comparable downstream performance while substantially improving ID purity, particularly under severe and dynamically varying OOD contamination.
\end{itemize}

\section{Related work}\label{sec:related}

\textbf{Active learning.} Pool-based AL balances \emph{informativeness} and \emph{diversity}. Uncertainty-based strategies rank samples by entropy~\citep{shannon1948mathematical}, margin~\citep{joshi2009multi,monarch2021human}, or Bayesian disagreement (BALD)~\citep{gal2017deep}; diversity-based methods add representativeness through core-set distances~\citep{sener2018active} or submodular optimization~\citep{kothawade2021similar}; hybrids such as BADGE~\citep{ashdeep} combine both~\citep{roy2001toward,tang2019self}; and generation-based approaches synthesize samples near the decision boundary~\citep{zhu2017generative}. All assume a centralized, \emph{closed-set} pool in which every unlabeled image is ID, critical classes are sufficiently represented, and the class distribution is balanced enough to avoid many redundant images. Clinical data violate all three assumptions: protocol and scanner changes create OOD images and domain shifts, rare diseases form long-tailed classes, and routine cases saturate the pool with redundant majority-class scans.

\textbf{Federated active learning.} Distributing training across privacy-sensitive clients amplifies these issues: local uncertainties are miscalibrated under non-IID data, and client-wise querying duplicates labels. Early prototypes used purely local scores~\citep{goetz2019active,ahn2024federated}, whereas \citet{Wu2022FederatedAL} scored each client's pool with the global model, and LoGo~\citep{10205184} couples local heterogeneity and global class imbalance through two-stage clustering. In federated medical imaging, FEAL~\citep{chen2024think} replaces softmax confidence with Dirichlet-based evidential learning that disentangles aleatoric from epistemic uncertainty under domain shift, FedAL~\citep{DENG2025303} applies ensemble-entropy querying over paired local and global models for skin lesions but remains purely uncertainty-driven and closed-set, and FAST~\citep{li2025fast} uses FM pseudo-labels to reduce communication and annotation rounds. All still overlook the open-set nature of medical data, allowing OOD samples to consume the budget while under-represented ID classes remain insufficiently queried~\citep{yu2023turning}.

\textbf{Open-set active learning.} OAL filters irrelevant OOD samples before labeling. LfOSA~\citep{ning2022active} detects known classes with Gaussian mixture models on activations, MQNet~\citep{park2022meta} and EOAL~\citep{safaei2024eoal} balance purity against informativeness with meta-learning or dual-entropy scores, and PAL~\citep{yang2023not} shows that selectively querying certain OOD images tightens ID decision boundaries. FocAL~\citep{SCHMIDT2024103162} adds a density-based OOD term to an uncertainty-decomposed histopathology score, but is centralized, validated on a single prostate cohort, and optimizes accuracy rather than annotation purity; FedPPD~\citep{bhatt2024federated} and SAFER~\citep{shen2025safer} target federated optimization or tabular electronic health records rather than image acquisition. A second wave removes the auxiliary detector altogether through single-stage joint scores (BUAL~\citep{zong2024bidirectional}, Contrastive OAL~\citep{yan2024contrastive}), a single Dirichlet-calibrated head (E2OAL~\citep{zong2026revisiting}), or joint OOD filtering and selection in a shared feature space (JODA~\citep{11092357}); a complementary line exploits vision-language priors, with CLIPNAL~\citep{heo2024clipnal} repurposing CLIPN's ``no'' prompt and OpenPath~\citep{zhong2025openpath} using GPT-4-generated OOD prompts for skin-lesion and histopathology data. All of these are studied in centralized settings and presuppose a globally observable OOD manifold, a shared pre-trained gate, or a fixed-threshold detector calibrated on the full pool, assumptions that break in federated workflows where OOD type, prevalence, and pre-training coverage vary across institutions and no client observes more than its local distribution.

FIDAL departs from these mechanisms in three ways: it operates strictly within the federated learning (FL) communication budget and never shares OOD samples, prompts, or pool-wide statistics; instead of a single pre-trained prior defining ``unknown'', it uses a per-client similarity filter against locally discovered OOD exemplars complemented by an FM-coverage term, so the gate adapts to each institution's artifact mix; and its rejection threshold is adaptive and tracks each client's evolving OOD distribution across rounds. Together, these choices make FIDAL viable in heterogeneous federations with under-represented classes, redundant majority samples, and drifting local OOD ratios.

\section{Methodology}\label{sec:method}

\begin{figure*}[!t]
    \centering
    \includegraphics[width=1.0\linewidth]{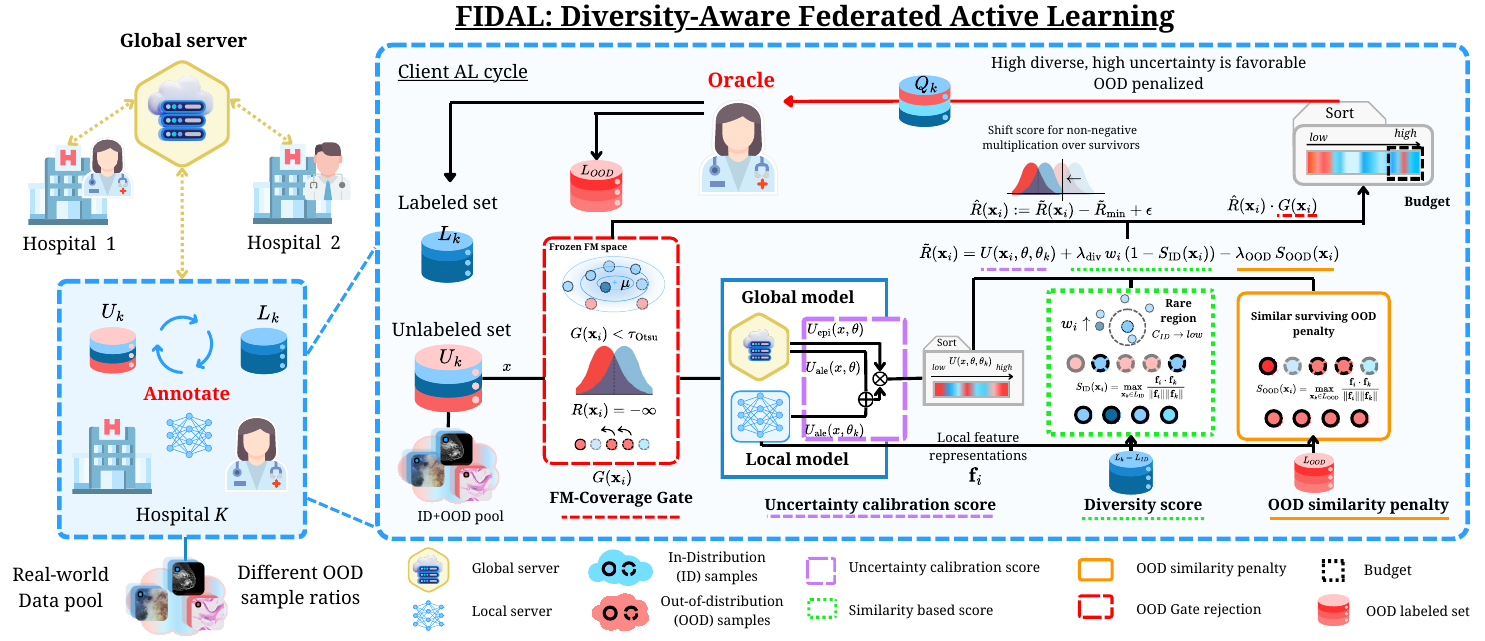}
    \caption{Overview of FIDAL. Calibrated uncertainty, diversity-aware sampling, and explicit OOD rejection are unified within a federated learning framework, enabling multi-center medical imaging models to approach centralized performance while reducing wasteful annotations on OOD artifacts.}
    \label{fig:method_proposal}
\end{figure*}

In a realistic OS-FAL context we define an \textbf{informative} sample as an unlabeled data point that, once labeled, is expected to maximize collective knowledge and robustness for both known and unknown classes across clients. FIDAL combines four mechanisms in a single ranking score: (i) a calibrated global--local evidential uncertainty (Section~\ref{subsec:uncertainty}); (ii) a similarity-based diversity term weighted by a support-set count $C_{ID}$ that promotes coverage of under-represented ID regions (Section~\ref{subsec:similarity}); (iii) an adaptive, per-client OOD penalty over the labeled-OOD similarity (Section~\ref{subsec:similarity}); and (iv) an FM Gaussian-coverage signal $G(\mathbf{x})$ with an adaptive Otsu rejection gate (Section~\ref{subsec:coverage}). The full scoring function is given by Eq.~\eqref{eq:coverage_fused}.

\subsection{Problem formulation}
Given an unlabeled dataset $\mathcal{U}$ and a labeled dataset $\mathcal{L}$, our goal is to construct a query set $\mathcal{Q}$ of size $B$ that (i) maximizes the inclusion of ID samples, (ii) minimizes the selection of OOD samples, and (iii) yields a diverse and informative labeled dataset. As illustrated in Fig.~\ref{fig:method_proposal}, $K$ clients each maintain a local model $\{\boldsymbol{\theta}_k\}_{k=1}^K$, while a global model $\boldsymbol{\theta}$ resides on the central server. Each client holds a labeled dataset $L_k$, an OOD-discovery labeled set $L_{k,OOD}$ (optional in the first AL round), and an unlabeled dataset $U_k = U_{k,ID}\cup U_{k,OOD}$. The OS-FAL process alternates federated training and local annotation. Initially, each client selects $B_k$ samples from $U_k$, annotates ID and OOD samples, updates $L_k^1$ and $L_{k,OOD}^1$, and revises the unlabeled set to $U_k^1 = U_k \setminus (L_k^1 \cup L_{k,OOD}^1)$. In each subsequent round $r$, a query set $Q_k^r$ is formed by the selection strategy below and $L_k^r$ and $U_k^r$ are updated accordingly; the process continues for $R$ rounds. To reduce notation, we present the strategy from the perspective of a single client.

\subsection{Uncertainty calibration}
\label{subsec:uncertainty}
Standard uncertainty metrics such as softmax entropy are miscalibrated under the domain shifts inherent to federated settings~\citep{chen2024think}. Following \emph{FEAL}, we adopt an evidential deep learning approach: given an input $\mathbf{x}$, the global model $\boldsymbol{\theta}$, and a local model $\boldsymbol{\theta}_k$, FIDAL computes the aleatoric uncertainty at both levels and uses the epistemic uncertainty of the more general global model to modulate their sum, calibrating the score against model ignorance caused by domain shift. Aleatoric uncertainty ($U_{\mathrm{ale}}$), representing inherent data ambiguity or noise, is
\begin{equation}\label{eq:aleatoric_feal_calib}
U_{\mathrm{ale}}(\mathbf{x},\boldsymbol{\theta}) = \sum_{c=1}^C \frac{\alpha_c}{S}\bigl[\psi(S+1)-\psi(\alpha_c+1)\bigr],
\end{equation}
and epistemic uncertainty ($U_{\mathrm{epi}}$), representing model ignorance due to lack of knowledge or domain shift, is
\begin{equation}\label{eq:epistemic_feal_calib}
\small
U_{\mathrm{epi}}(\mathbf{x},\boldsymbol{\theta}) = \sum_{c=1}^C\left[ \log\Gamma(\alpha_c)-\log\Gamma(S) - (\alpha_c-1)\bigl[\psi(\alpha_c)-\psi(S)\bigr] \right].
\end{equation}
Here, $\boldsymbol{\alpha} = \mathbf{e} + 1$ are the Dirichlet parameters derived from the evidence $\mathbf{e} = \text{activation}(f(\mathbf{x};\boldsymbol{\theta}))$, $C$ is the number of classes, $S=\sum_{c=1}^C\alpha_c$ is the Dirichlet strength with per-class parameter $\alpha_c$, and $\psi$ and $\Gamma$ are the digamma and Gamma functions~\citep{chen2024think}. We incorporate aleatoric uncertainty from both models to capture inherent ambiguity and client noise, but rely on global epistemic uncertainty alone to detect domain shift, avoiding the overconfidence of local models trained on narrow distributions. The calibrated uncertainty is
\begin{equation} \label{eq:fidal_uncertainty_calib}
U(\mathbf{x},\boldsymbol{\theta},\boldsymbol{\theta}_k) = \bigl[U_{\mathrm{ale}}(\mathbf{x},\boldsymbol{\theta})+U_{\mathrm{ale}}(\mathbf{x},\boldsymbol{\theta}_k)\bigr] \cdot U_{\mathrm{epi}}(\mathbf{x},\boldsymbol{\theta}).
\end{equation}

\subsection{Similarity-based diversity, OOD penalty, and ranking score}\label{subsec:similarity}
To prioritize the most relevant ID samples, we use a similarity-based scoring mechanism that accounts for both diversity and OOD detection. For each unlabeled candidate $\mathbf{x}_i$ we measure its cosine similarity to the labeled ID set,
\begin{equation}\label{eq:sid}
S_{\text{ID}}(\mathbf{x}_i) = \max_{\mathbf{x}_j \in L_{\text{ID}}} \frac{\mathbf{f}_i \cdot \mathbf{f}_j}{\|\mathbf{f}_i\| \|\mathbf{f}_j\|},
\end{equation}
where $\mathbf{f}_i = h(\mathbf{x}_i)$ is the penultimate-layer embedding of $\mathbf{x}_i$ under the client's local model $\boldsymbol{\theta}_k$ (distinct from the frozen FM encoder $g(\cdot)$ of Section~\ref{subsec:coverage}, which is used only for the coverage signal) and $L_{\text{ID}}$ is the client's labeled ID data. Analogously, the OOD penalty is the similarity to the client's labeled OOD data $L_{\text{OOD}}$:
\begin{equation}\label{eq:sood}
S_{\text{OOD}}(\mathbf{x}_i) = \max_{\mathbf{x}_m \in L_{\text{OOD}}} \frac{\mathbf{f}_i \cdot \mathbf{f}_m}{\|\mathbf{f}_i\| \|\mathbf{f}_m\|}.
\end{equation}
To capture local density and target under-represented regions of the ID feature space, we define a support-set count of the labeled and unlabeled points that sit next to the candidate,
\begin{equation}\label{eq:cid}
C_{ID}(\mathbf{x}_i) = \sum_{{\mathbf{x}_j \in  (U \cup L_{\text{ID}})}} \mathds{1}\{S(\mathbf{x}_i, \mathbf{x}_j) > \tau_{ID}\},
\end{equation}
where $S(\mathbf{x}_i, \mathbf{x}_j)$ is the same cosine similarity used in $S_{\text{ID}}$ and $\tau_{ID} = 0.8\cdot S_{\text{ID}}(\mathbf{x}_i)$ is an adaptive threshold, so that $C_{ID}$ counts the neighbors whose similarity lies within $20\%$ of the candidate's maximum ID similarity. A large $C_{ID}$ suggests a densely sampled region where additional labels may be redundant; a small one indicates a rare region worth exploring. The computation is a single vectorized matrix multiplication; despite a theoretical complexity of $\mathcal{O}(|U| \cdot N \cdot d)$ with $N=|U \cup L|$, its runtime is negligible on modern accelerators. Combining uncertainty, support-weighted diversity, and OOD rejection yields the task-feature ranking score
\begin{equation}\label{eq:softmax_def_calib}
   \tilde{R}(\mathbf{x}_i)
  = U(\mathbf{x}_i,\boldsymbol{\theta},\boldsymbol{\theta}_k)
    + \lambda_{\mathrm{div}}\,w_i\,(1 - S_{ID}(\mathbf{x}_i))
    - \lambda_{\mathrm{OOD}}\,S_{\mathrm{OOD}}(\mathbf{x}_i),
\end{equation}
where $\lambda_{\text{div}}$ and $\lambda_{\text{OOD}}$ weigh diversity and OOD rejection, and $w_i = \exp(-C_{ID}(\mathbf{x}_i)) / \sum_j \exp(-C_{ID}(\mathbf{x}_j))$ is a softmax over the negative counts.

\subsection{Foundation-model coverage and adaptive gate}\label{subsec:coverage}
Finally, FIDAL carries the Gaussian-RBF ``uncertainty coverage'' idea of Uncertainty Herding~\citep{3e1026bf53124ed3baa70ebbdb63cfc4} into the federated, OOD-aware setting. Each candidate is scored by its proximity to the labeled-ID manifold under a per-client diagonal Gaussian whose moments $(\boldsymbol{\mu}, \boldsymbol{\sigma}^2)$ are estimated directly from $L_{ID}$ in frozen FM space, yielding a single one-dimensional coverage score $G(\mathbf{x})$ that fits naturally with the adaptive thresholding step below. Let $g(\cdot)$ be a frozen FM encoder (PanDerm, CHIEF, or MammoCLIP) producing $\mathbf{z}_i = g(\mathbf{x}_i) \in \mathbb{R}^d$. For each client we fit
\begin{equation}\label{eq:moments}
\boldsymbol{\mu} = \frac{1}{|L_{ID}|} \sum_{\mathbf{x}_j \in L_{ID}} \mathbf{z}_j,\qquad
\boldsymbol{\sigma}^2 = \frac{1}{|L_{ID}|-1} \sum_{\mathbf{x}_j \in L_{ID}} (\mathbf{z}_j - \boldsymbol{\mu})^2,
\end{equation}
with per-coordinate variance $\boldsymbol{\sigma}^2$, and compute the diagonal Gaussian log-likelihood
\begin{equation}\label{eq:fm_likelihood}
\log G(\mathbf{x}_i) = -\tfrac{1}{2}\sum_{c=1}^{d}\!\Big[\tfrac{(z_{i,c} - \mu_c)^2}{\sigma_c^2} + \log \sigma_c^2\Big] + \text{const}.
\end{equation}
Throughout, $G(\mathbf{x})$ denotes this log-likelihood after min--max normalization over the unlabeled pool to $[0,1]$; it is a per-client, task-agnostic estimate of how well an FM-encoded candidate aligns with the labeled-ID manifold. Up to the diagonal-covariance assumption and an additive constant, Eq.~\eqref{eq:fm_likelihood} is the log of an axis-aligned anisotropic Gaussian RBF centered at $\boldsymbol{\mu}$ with per-coordinate bandwidth $\sigma_c$, so $G(\mathbf{x})$ preserves the Gaussian-kernel intuition that motivates Uncertainty Herding.

\paragraph{Adaptive gating on the FM.} Having reduced the coverage signal to a one-dimensional distribution $\{G(\mathbf{x}_i)\}_{i \in U_k}$, we no longer need a hand-tuned percentile cut-off. Instead, we apply Otsu's between-class variance criterion~\citep{4310076} to this distribution, yielding a per-client, per-round threshold $\tau_{\text{Otsu}}$ that adapts to each client's OOD prevalence and to its evolution across rounds. Candidates with $G(\mathbf{x}_i) < \tau_{\text{Otsu}}$ are hard-rejected; survivors $S_k = \{i \in U_k : G(\mathbf{x}_i) \geq \tau_{\text{Otsu}}\}$ retain $G(\mathbf{x}_i)$ as a soft multiplicative coverage weight. To preserve rank monotonicity in the negative tail of $\tilde{R}(\mathbf{x}_i)$ from Eq.~\eqref{eq:softmax_def_calib}, we shift the base score to be non-negative over survivors before multiplication:
\begin{equation}\label{eq:coverage_fused}
R(\mathbf{x}_i) =
\begin{cases}
\bigl(\tilde{R}(\mathbf{x}_i) - \min_{j \in S_k} \tilde{R}(\mathbf{x}_j) + \epsilon\bigr) \cdot G(\mathbf{x}_i), & i \in S_k, \\
-\infty, & i \notin S_k.
\end{cases}
\end{equation}
The query set $\mathcal{Q}$ consists of the $B$ unlabeled samples with the largest $R(\mathbf{x}_i)$. This scoring decouples \emph{what counts as OOD} from \emph{how survivors are ranked}: the gate inherits the robustness of the frozen FM prior, while the ranking still benefits from task-fine-tuned uncertainty and similarity signals.

\section{Experiments and results}\label{sec:experiments}

\subsection{Datasets and open-set protocol}
\label{subsec:datasets}
We evaluate FIDAL on three multi-center medical imaging benchmarks with inherent domain shifts. Train/test splits are dataset-specific: FedISIC uses the FLamby split, FedCamelyon an $8{:}2$ patient-level split, and FedEMBED the base split of \citet{ROSCHEWITZ2025103668} ($\approx$$77{:}23$). To probe OOD robustness, we inject OOD samples into each client's \emph{unlabeled} pool at a high prevalence for FedISIC ($38.7\%$) and FedCamelyon ($38.5\%$), preserving the original class proportions, whereas FedEMBED carries an organically occurring OOD prevalence. Per-client OOD compositions are listed in Supplementary Tables~S1 and~S2.
\begin{itemize}
    \item \textbf{FedISIC}~\citep{chen2024think,ogier2022flamby}: a public multi-center skin-lesion benchmark comprising the four largest FLamby sites, with $9{,}930$, $3{,}163$, $2{,}691$ and $1{,}807$ training images across eight classes (MEL, NV, BCC, AK, BKL, DF, VASC, SCC). OOD samples are clinical non-dermoscopy photographs from DDI (all $656$ images)~\citep{daneshjou2022disparities} and a subsample of $10{,}468$ of the $16{,}577$ Fitzpatrick17k images~\citep{groh2021evaluating}, which capture clinically relevant shifts in imaging conditions and skin tones and are visually distinct from the dermoscopy ID data.
    \item \textbf{FedCamelyon}~\citep{bandi2018detection,chen2024think}: histology patches from five centers for binary tumor classification, with $12{,}360$, $7{,}256$, $17{,}687$, $27{,}001$ and $30{,}513$ training samples (ID + OOD) per center, i.e., $58{,}348$ ID and $36{,}469$ OOD across the federation. OOD patches come exclusively from HistoArtifacts~\citep{kanwal2024equipping} at 40$\times$ magnification and span four slide artifacts: blood ($22{,}738$), blur ($6{,}797$), bubbles ($3{,}344$) and tissue damage ($3{,}590$), stratified across the five clients.
    \item \textbf{FedEMBED}: breast-density classification (BI-RADS A--D) from the EMBED dataset~\citep{jeong2023emory}, acquired in four hospitals with heterogeneous devices. Following the base splits of \citet{ROSCHEWITZ2025103668}, $222{,}700$ training and $65{,}891$ test images are distributed across four scanner clients: Selenia Dimensions ($198{,}726$ / $58{,}925$), Senograph 2000D ADS ($10{,}043$ / $2{,}810$), Lorad Selenia ($7{,}749$ / $2{,}421$), and Clearview CSm ($6{,}182$ / $1{,}735$) (train / test). As real-world OOD we use the five expert-annotated artifact types of \citet{10981259} (circular and triangular skin markers, breast implants, support devices, and spot-compression structures), which originate from the same scanners, occur naturally in deployment, and have been shown to degrade model performance; they amount to $\approx$$13\%$ of the public EMBED data, with prevalence varying from $4.8\%$ to $13.1\%$ across clients (Supplementary Table~S2).
\end{itemize}

\subsection{Experimental setup}\label{subsec:baselines}
\textbf{Baselines.} We compare \texttt{FIDAL} with four standard AL methods adapted to the federated setting, \texttt{Random}, \texttt{Entropy}~\citep{shannon1948mathematical}, \texttt{BADGE}~\citep{ashdeep}, and the federated evidential method \texttt{FEAL}~\citep{chen2024think}, and with four open-set methods: the GMM-based \texttt{LfOSA}~\citep{ning2022active}, \texttt{EOAL}~\citep{safaei2024eoal}, \texttt{PAL}~\citep{yang2023not}, and the vision-language-gated \texttt{OpenPath}~\citep{zhong2025openpath}, run federated as \texttt{OpenPath*} (each client gates its own pool with a frozen BiomedCLIP encoder against GPT-4-generated prompts, sharing neither prompts nor embeddings; Supplementary Section~S2). A fully supervised model trained on the clean ID pool marks the ceiling.

\textbf{Metrics.} We report balanced multi-class accuracy (BMA) on FedISIC and FedEMBED (supervised references $67.71\%$ and $69.54\%$) and accuracy (ACC) on binary FedCamelyon ($97.72\%$), together with the \emph{ID purity}, the share of the cumulative labeled pool that is ID at a given round. Because the pool includes the shared random warm-up round, purity is directly comparable across methods. Unless stated otherwise, each method is reported at its own best round (mean$\pm$std over 3 seeds), so that early peakers are not penalized, and differences are given in percentage points (pp).

\textbf{Protocol.} We run $R{=}5$ rounds on FedISIC and $R{=}10$ on FedCamelyon and FedEMBED, with $E{=}100$ local epochs per round ($E{=}50$ for FedEMBED), $K{=}4$ clients ($K{=}5$ for FedCamelyon), and a per-round budget $B_k{=}500$. Backbones are EfficientNet-B0~\citep{tan2019efficientnet}, DenseNet-121~\citep{huang2017densely}, and a linear probe on a frozen MammoCLIP encoder~\citep{ghosh2024mammo}, respectively, with the public preprocessing of each benchmark~\citep{ogier2022flamby,Jiang_Wang_Dou_2022,ghosh2024mammo}; optimizer settings are listed in Supplementary Section~S2. \texttt{FIDAL}'s coverage encoder $g(\cdot)$ is PanDerm~\citep{yan2025multimodal}, CHIEF~\citep{wang2024pathology}, or MammoCLIP~\citep{ghosh2024mammo}, and its only selection hyperparameters, $(\lambda_{\text{div}},\lambda_{\text{OOD}})$, are set per dataset from the sensitivity analysis of Section~\ref{subsec:hyperparam}: $(0.5,0.5)$, $(1.0,1.0)$, and $(0.5,5.0)$ for FedISIC, FedCamelyon, and FedEMBED.

\begin{figure*}[!t]
    \centering
    \includegraphics[width=0.8\linewidth]{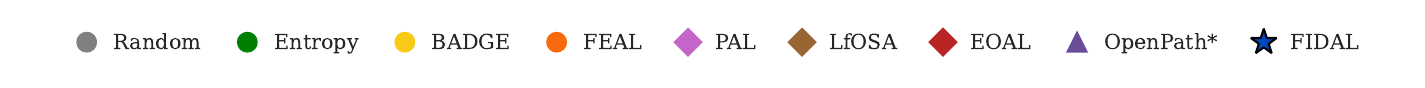}\\
    \includegraphics[width=\linewidth]{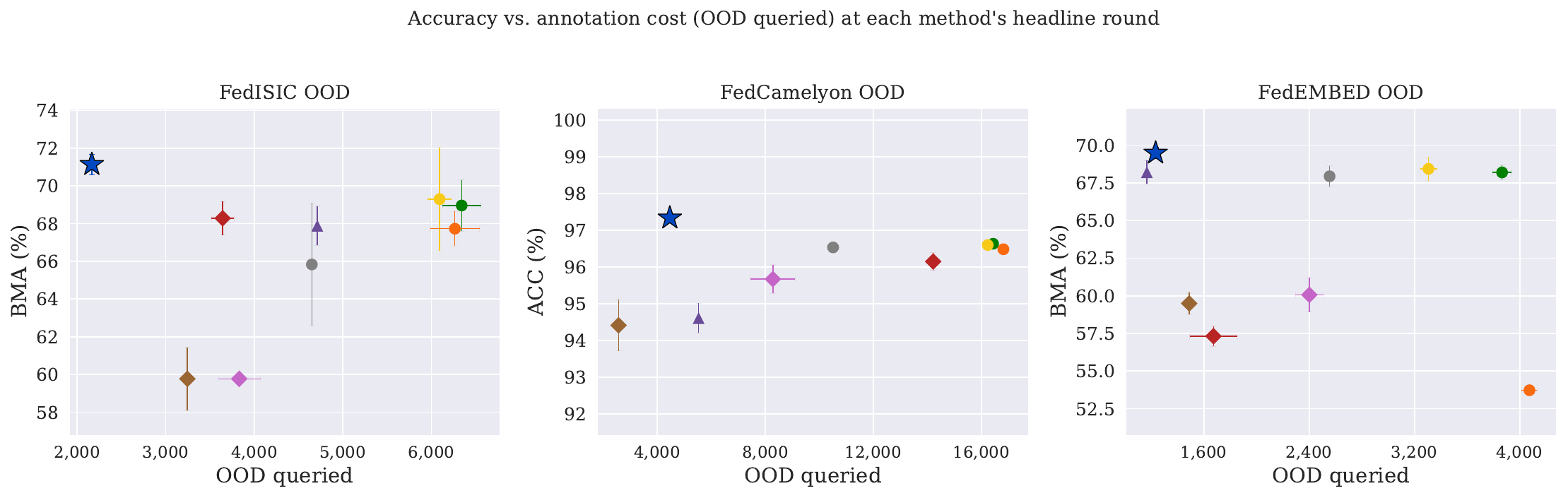}
    \caption{\textbf{Accuracy versus OOD queried.} Each method is placed at its headline operating point (Table~\ref{tab:headline_3datasets}): the classification metric it attains (BMA for FedISIC/FedEMBED, ACC for FedCamelyon) against the cumulative number of OOD samples it sent for annotation to reach that round (fewer is better), summed over clients and averaged over 3 seeds; error bars are $\pm 1$ std. \texttt{FIDAL} (blue star) occupies the top-left corner of every dataset. Because methods peak at different rounds, the plotted counts span different numbers of query rounds and are not directly comparable with the matched-round counts of Section~\ref{subsec:ood_cost}.}
    \label{fig:bma_vs_oodqueried}
\end{figure*}

\subsection{Results across the three benchmarks}\label{subsec:ood_cost}\label{subsec:per_benchmark}
\begin{figure*}[!t]
    \centering
    \includegraphics[width=\linewidth]{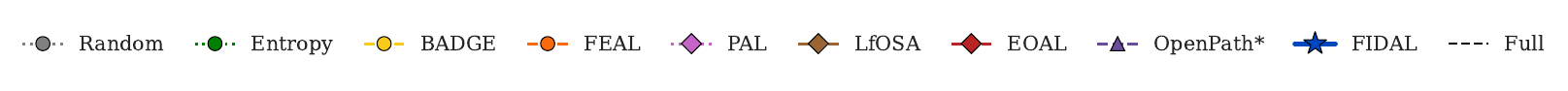}\\[-0.2em]
    \includegraphics[width=0.82\linewidth]{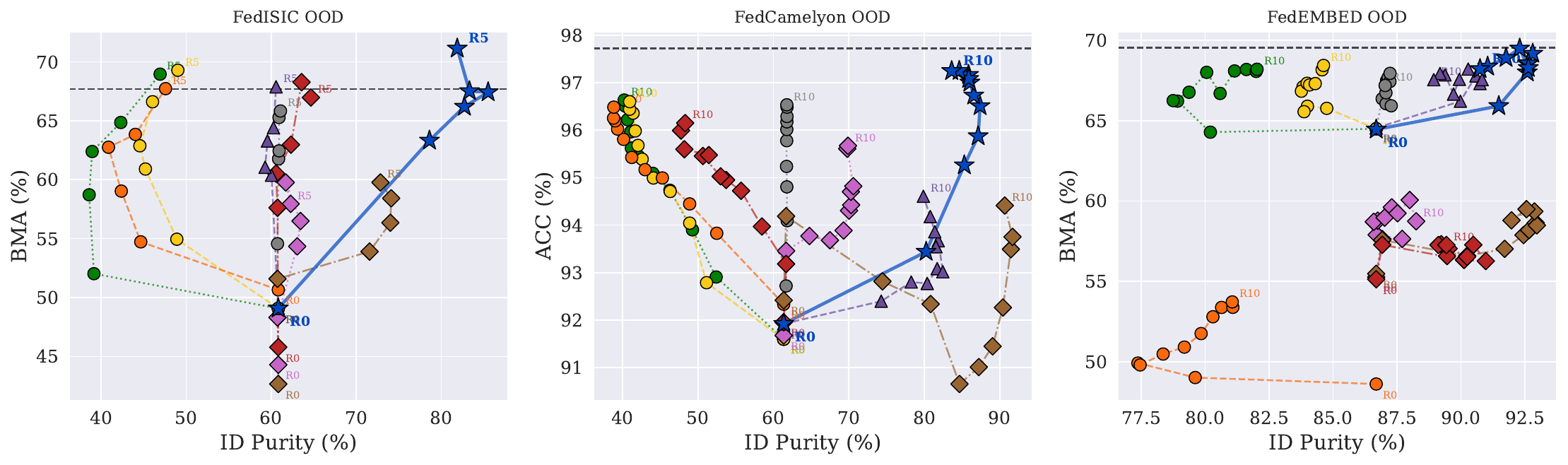}
    \caption{\textbf{Pareto view: classification metric versus ID purity.} Each path is one method's federation-wide (ID purity, BMA/ACC) trajectory across AL rounds $0\to R_{\max}$ ($R_{\max}{=}5$ for FedISIC, $10$ otherwise), one panel per dataset; top-right is best. The same view decomposed by client is given in Supplementary Fig.~S1.}
    \label{fig:pareto_bma_idp_3datasets_global}
\end{figure*}

\textbf{Best accuracy and best purity at once.} Table~\ref{tab:headline_3datasets} and Fig.~\ref{fig:bma_vs_oodqueried} summarize the headline comparison at each method's best round. \texttt{FIDAL} reaches $71.14 \pm 0.54\%$ BMA at $81.9\%$ ID purity on FedISIC, $97.34 \pm 0.12\%$ ACC at $83.8\%$ on FedCamelyon, and $69.49 \pm 0.36\%$ BMA at $92.3\%$ on FedEMBED. It is the closest method to the fully supervised reference on every dataset, exceeding it on FedISIC and matching it on FedEMBED, and the only method on the accuracy--ID purity Pareto front of all three (Fig.~\ref{fig:pareto_bma_idp_3datasets_global}); the per-client trajectories (Supplementary Fig.~S1) show that no single center drives the gain. Every baseline sits on one side of a trade-off. The uncertainty methods (\texttt{Entropy}, \texttt{BADGE}) come within $1.85$\,pp of \texttt{FIDAL} on FedISIC and $0.71$\,pp on FedCamelyon but spend roughly half of their budget on OOD ($47$--$49\%$ and $40$--$41\%$ ID purity). The detector-based open-set methods buy purity with accuracy: \texttt{LfOSA} reaches $72.9$, $90.7$, and $92.6\%$ purity on the three datasets but trails \texttt{FIDAL} by $11.4$, $2.93$, and $10.0$\,pp, and \texttt{PAL}, \texttt{LfOSA}, and \texttt{EOAL} collapse to $57$--$60\%$ BMA on FedEMBED ($9$--$12$\,pp behind). The vision-language gate of \texttt{OpenPath*} holds accuracy on FedEMBED ($68.21\%$) but not on FedCamelyon ($94.61\%$).

\begin{table*}[!t]
\centering
\caption{Headline three-dataset comparison: classification metric (BMA for FedISIC and FedEMBED, ACC for FedCamelyon binary classification; \%, mean$\pm$std across 3 seeds) at each method's best AL round (in parentheses) and the cumulative ID purity (\%) at that round. FedISIC $B{=}500$/$E{=}100$ ($R\le 5$); FedCamelyon $B{=}500$/$E{=}100$ ($R\le 10$); FedEMBED $B{=}500$/$E{=}50$ ($R\le 10$). Best per-column score / ID purity in \textbf{bold} / \underline{underlined}.}
\label{tab:headline_3datasets}
\resizebox{\textwidth}{!}{%
\begin{tabular}{l @{\hspace{10pt}} cc @{\hspace{10pt}} cc @{\hspace{10pt}} cc @{\hspace{10pt}} cc}
\toprule
\multirow{2}{*}{\textbf{Method}}
  & \multicolumn{2}{c}{\textbf{FedISIC} (OOD)}
  & \multicolumn{2}{c}{\textbf{FedCamelyon} (OOD)}
  & \multicolumn{2}{c}{\textbf{FedEMBED} (OOD)}
  & \multicolumn{2}{c}{\textbf{Avg.}} \\
\cmidrule(lr){2-3}\cmidrule(lr){4-5}\cmidrule(lr){6-7}\cmidrule(lr){8-9}
 & BMA & ID Purity & ACC & ID Purity & BMA & ID Purity & Metric & ID Purity \\
\midrule
Random                                  & 65.83\,$\pm$2.66~(R5) & 61.1 & 96.53\,$\pm$0.13~(R10) & 61.8 & 67.94\,$\pm$0.57~(R9)  & 87.2 & 76.77 & 70.0 \\
Entropy~\citep{shannon1948mathematical}  & 68.96\,$\pm$1.11~(R5) & 46.9 & 96.63\,$\pm$0.04~(R10) & 40.2 & 68.19\,$\pm$0.40~(R10) & 82.0 & 77.93 & 56.4 \\
BADGE~\citep{ashdeep}                    & 69.29\,$\pm$2.23~(R5) & 49.0 & 96.60\,$\pm$0.09~(R10) & 40.9 & 68.43\,$\pm$0.64~(R10) & 84.6 & 78.11 & 58.2 \\
FEAL~\citep{chen2024think}               & 67.73\,$\pm$0.75~(R5) & 47.6 & 96.49\,$\pm$0.09~(R10) & 38.8 & 53.72\,$\pm$0.21~(R10) & 81.1 & 72.65 & 55.8 \\
\midrule
PAL~\citep{yang2023not}                  & 59.77\,$\pm$0.18~(R4) & 61.7 & 95.67\,$\pm$0.39~(R10) & 69.9 & 60.06\,$\pm$0.94~(R9)  & 88.0 & 71.83 & 73.2 \\
LfOSA~\citep{ning2022active}             & 59.77\,$\pm$1.37~(R5) & 72.9 & 94.41\,$\pm$0.70~(R10) & \underline{90.7} & 59.49\,$\pm$0.60~(R9)  & \underline{92.6} & 71.22 & 85.4 \\
EOAL~\citep{safaei2024eoal}              & 68.28\,$\pm$0.73~(R4) & 63.6 & 96.15\,$\pm$0.24~(R10) & 48.3 & 57.31\,$\pm$0.55~(R7)  & 89.2 & 73.91 & 67.0 \\
OpenPath*~\citep{zhong2025openpath}      & 67.89\,$\pm$0.85~(R5) & 60.6 & 94.61\,$\pm$0.41~(R10) & 79.9 & 68.21\,$\pm$0.63~(R5)  & 90.3 & 76.90 & 76.9 \\
\midrule
\textbf{FIDAL}                          & \textbf{71.14\,$\pm$0.54}~(R5) & \underline{81.9} & \textbf{97.34\,$\pm$0.12}~(R10) & 83.8 & \textbf{69.49\,$\pm$0.36}~(R7) & 92.3 & \textbf{79.32} & \underline{86.0} \\
\midrule
\textit{Fully supervised}    & \textit{67.71} & \textit{100} & \textit{97.72} & \textit{100} & \textit{69.54} & \textit{100} & \textit{78.32} & \textit{100} \\
\bottomrule
\end{tabular}}
\end{table*}

\textbf{What the purity buys.} We measure annotation cost as the cumulative number of OOD samples sent to the oracle, evaluated for every method at \texttt{FIDAL}'s headline round and compared only with accuracy-matched baselines, so that a low count bought by an accuracy collapse is not credited as a saving. \texttt{FIDAL} queries $\approx 2{,}165$ OOD samples on FedISIC versus $4{,}222$--$6{,}342$ for \texttt{BADGE}, \texttt{Entropy}, and \texttt{EOAL} ($2.0$--$2.9\times$ fewer); $\approx 4{,}465$ OOD patches on FedCamelyon versus $10{,}512$--$16{,}819$ for \texttt{Random}, \texttt{Entropy}, \texttt{BADGE}, and \texttt{FEAL} ($2.4$--$3.8\times$ fewer); and $\approx 1{,}233$ on FedEMBED versus $1{,}647$--$3{,}017$ for \texttt{Random}, \texttt{Entropy}, \texttt{BADGE}, and \texttt{OpenPath*} ($1.3$--$2.4\times$ fewer). Only \texttt{LfOSA} queries fewer OOD samples at these rounds (FedCamelyon $2{,}567$, FedEMBED $1{,}126$), at the cost of $2.93$ and $10.00$\,pp. The round-by-round composition of the labeled pool (Supplementary Figs.~S2--S4) shows where the difference arises: \texttt{FIDAL} keeps cumulative OOD below $22\%$ on FedISIC and near $9\%$ on FedEMBED, whereas \texttt{Entropy} spends up to $61\%$ of an early-round budget on OOD on FedISIC and $20\%$ on FedEMBED. Under a fixed budget every avoided OOD query is an extra ID label: relative to \texttt{Random}, \texttt{FIDAL} acquires $+715$ more ID samples on FedISIC by $R{=}1$ and $+2{,}240$ by $R{=}4$.

\textbf{Robust to tight budgets and few local epochs.} The full budget and local-epoch sweeps (Supplementary Tables~S3--S5) confirm the headline pattern. On FedISIC, \texttt{FIDAL} tops every cell, with BMA rising from $50.71\%$ at $B{=}50$ to $71.14\%$ at $B{=}500$ and from $61.47\%$ at $E{=}2$ to $71.14\%$ at $E{=}100$; the largest margins appear where the local model is weakest. It loses only $9.7$\,pp at $E{=}2$, against $35$\,pp for \texttt{FEAL}, $30$\,pp for \texttt{EOAL}, and $21$--$24$\,pp for \texttt{PAL} and \texttt{LfOSA}. On FedCamelyon it is best in every cell but the tightest ($B{=}50$, where \texttt{FEAL} leads by $0.4$\,pp at $51.2\%$ purity versus $96.1\%$) and averages $94.35\%$ ACC and $89.8\%$ purity over the seven cells, while no other method exceeds $73.8\%$ purity. On FedEMBED, at the final round, it averages $66.88\%$ BMA against $66.73\%$ for \texttt{OpenPath*}, with the highest purity of any non-LfOSA method ($92.5\%$); the frozen-encoder linear probe is nearly epoch-invariant ($98\%$ of peak at $E{=}2$).

\begin{figure*}[p]
    \centering
    \includegraphics[width=\linewidth]{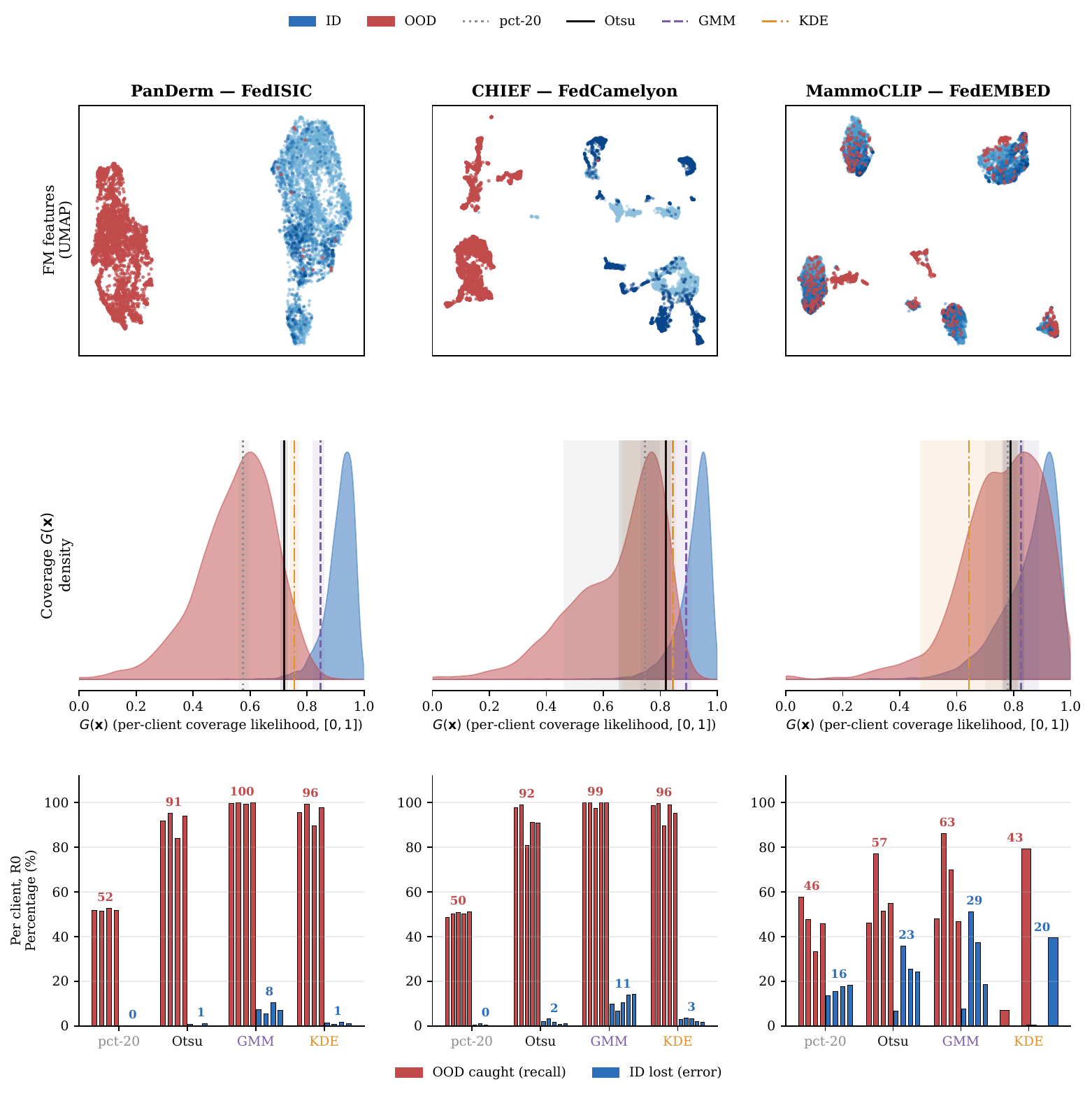}
    \caption{\textbf{How the FM-coverage gate works, per dataset.} \textbf{Top:} UMAP of the raw FM features, ID (blue, shaded by class) versus OOD (red); PanDerm and CHIEF separate them, MammoCLIP overlaps. \textbf{Middle:} the per-client coverage signal $G(\mathbf{x})$, ID (blue) versus OOD (red); for each rule the shaded band spans the per-client thresholds and the line is the median cut. \textbf{Bottom:} per-client $R{=}0$ comparison of four automatic gates (pct-20 / Otsu / GMM / KDE): OOD caught (recall, red) versus ID lost (error, blue), one bar per client, with the mean over clients annotated above each cluster. Otsu gives the best recall at minimal ID loss and reproduces the deployed per-client gate ($\approx$90\% OOD caught at $\le\!2\%$ ID loss on FedISIC/FedCamelyon).}
    \label{fig:fm_story}
\end{figure*}

\subsection{Why FIDAL works: the coverage gate and the ablation}\label{subsec:gate}\label{sec:ablation}
\textbf{The gate follows the foundation model's geometry.} Fig.~\ref{fig:fm_story} traces the FM-coverage gate on each dataset. In the UMAP of the raw FM features (top row), PanDerm and CHIEF push the OOD samples off the ID manifold, whereas MammoCLIP leaves the organic mammography artifacts mixed into it. The one-dimensional coverage signal $G(\mathbf{x})$ inherits this geometry (middle row): cleanly bimodal on FedISIC and FedCamelyon, overlapping on FedEMBED. The bottom row compares four automatic thresholds at the first round, one bar per client. A fixed bottom-$20\%$ percentile catches only $\approx$$50\%$ of the OOD on the $38\%$-OOD pools yet discards $16\%$ of the ID on low-prevalence FedEMBED; a two-component GMM reaches $\ge\!99\%$ recall on the separable datasets but over-rejects ID ($8$--$11\%$, and $29\%$ on FedEMBED); a KDE density valley tracks Otsu where the modes separate ($\approx$$96\%$) but is unstable where they overlap. Otsu catches $\approx$$91\%$ of the OOD at $\approx$$1\%$ ID loss on FedISIC and $\approx$$92\%$ at $\approx$$2\%$ on FedCamelyon, without any hand-tuned threshold. On FedEMBED no rule separates the overlapping modes (Otsu: $57\%$ recall at $23\%$ ID loss), so the gate hands the purity load to the $\lambda_{\text{OOD}}\,S_{\mathrm{OOD}}$ penalty. The two mechanisms are complementary: the gate rejects novel OOD types that the labeled-OOD set has not yet seen, and $S_{\mathrm{OOD}}$ catches samples that resemble artifacts already labeled.

\begin{figure*}[!t]
    \centering
    \includegraphics[width=\linewidth]{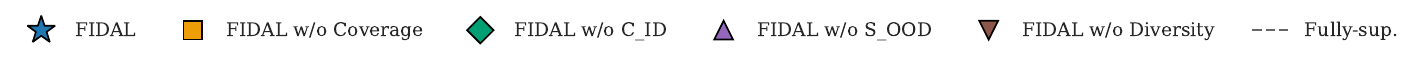}\\[-0.2em]
    \includegraphics[width=.8\linewidth]{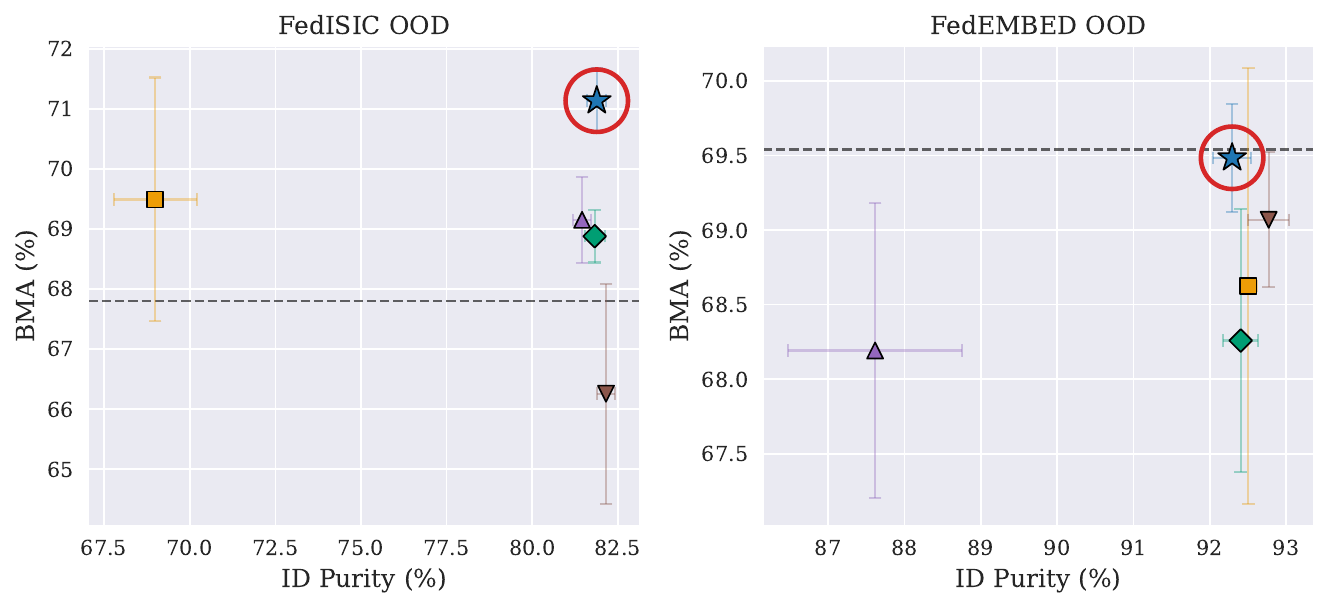}
    \caption{\textbf{Component ablation.} BMA versus cumulative ID purity ($\pm 1\sigma$ over 3 seeds; red circle: adopted \texttt{FIDAL}). \textbf{Left:} FedISIC OOD ($R{=}5$, base $(0.5,0.5)$): \texttt{FIDAL} is Pareto-optimal; removing the coverage gate drops ID purity $\approx$13\,pp, removing diversity costs $\approx$$4.9$\,pp BMA. \textbf{Right:} FedEMBED OOD ($R{=}7$, base $(0.5,5.0)$): \texttt{FIDAL} tops BMA ($69.49$\%) at near-top ID purity, but variants remain within $\approx$1\,pp at the label-noise ceiling; the one clear signal is \texttt{w/o $S_{\mathrm{OOD}}$} dropping ID purity $92.3{\to}87.6$\%, motivating $\lambda_{\text{OOD}}{=}5$.}
    \label{fig:fedisic_ablation}
\end{figure*}

\textbf{Each component owns one axis.} Removing one mechanism at a time (Fig.~\ref{fig:fedisic_ablation}) separates the two axes of the result. On FedISIC, dropping the coverage gate lowers ID purity by $\approx$$13$\,pp (to $69.0\%$) but costs only $1.65$\,pp BMA, so the gate is the purity mechanism; dropping the diversity term costs the most accuracy ($66.26\%$, $-4.88$\,pp) while retaining the highest purity, so a \emph{varied} set of ID samples, not OOD rejection alone, drives downstream accuracy. Removing the support-set weighting or the labeled-OOD penalty costs $2.0$--$2.3$\,pp each at matched purity, and only the full combination is Pareto-optimal ($71.14\%$ BMA at $81.9\%$ purity). On FedEMBED all variants stay within $\approx$$1$\,pp BMA at the label-noise ceiling, and the one clear signal is the labeled-OOD penalty: removing it drops purity from $92.3$ to $87.6\%$, which is why FedEMBED uses $\lambda_{\text{OOD}}{=}5$. Where the FM separates the OOD, the gate alone filters; where artifacts share the ID manifold, the explicit similarity penalty becomes the primary purity mechanism.

\section{Discussion}\label{sec:discussion}
While FIDAL builds on established components, its contribution lies in \emph{how} these are combined and adapted to the OS-FAL setting in medical imaging: a calibrated global--local evidential score, support-based diversity for rare-class coverage, and adaptive per-client OOD rejection. Across three benchmarks, this design achieves accuracy--purity trade-offs not matched by the evaluated baselines in the tested settings.

\subsection{Impact of FIDAL's budget efficiency in real-world heterogeneous settings}\label{subsec:disc_rare}\label{subsec:disc_artifacts}
\begin{figure}[!t]
    \centering
    \includegraphics[width=\linewidth]{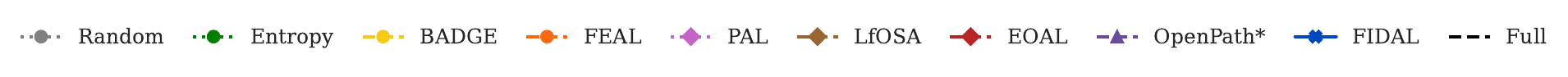}\\[-0.2em]
    \begin{tabular}{@{}cc@{}}
        \includegraphics[width=0.47\linewidth]{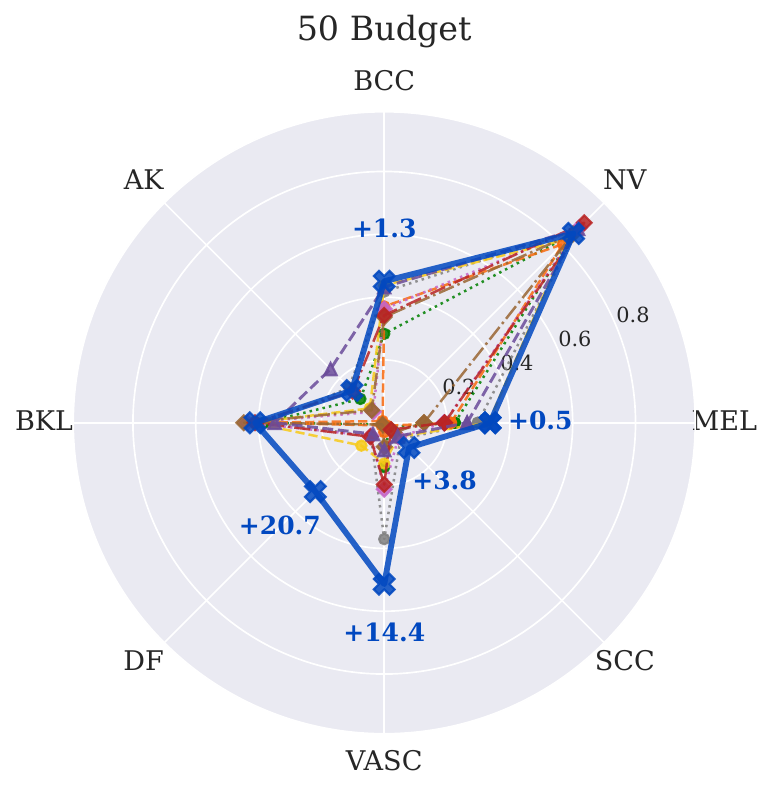} &
        \includegraphics[width=0.47\linewidth]{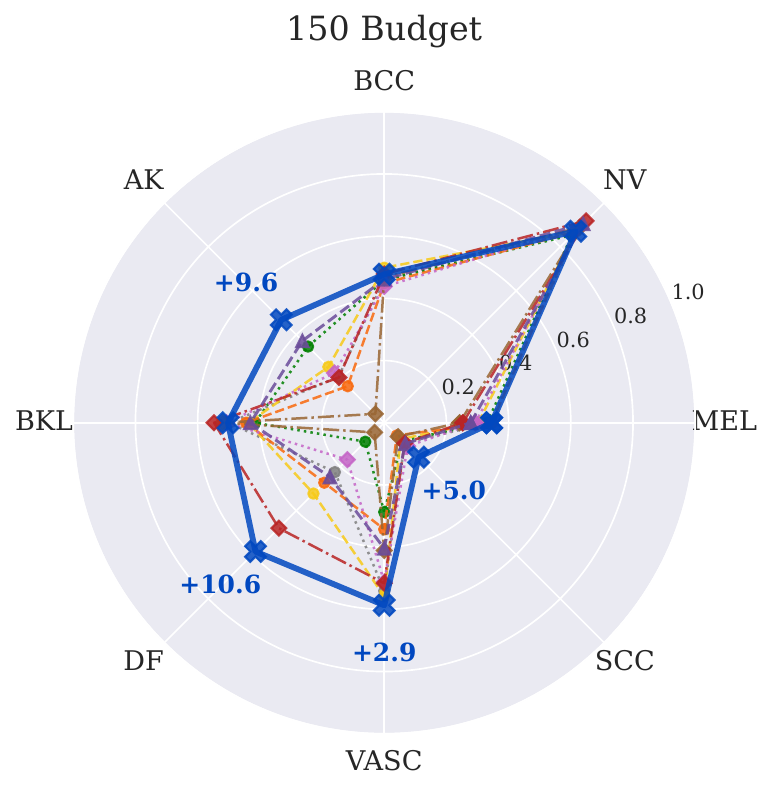} \\
        \includegraphics[width=0.47\linewidth]{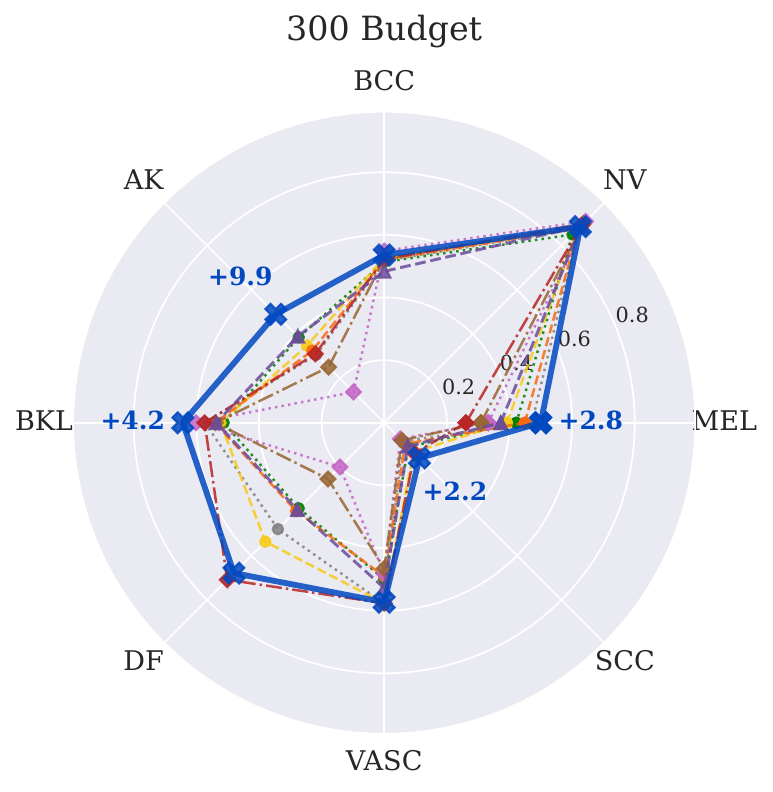} &
        \includegraphics[width=0.47\linewidth]{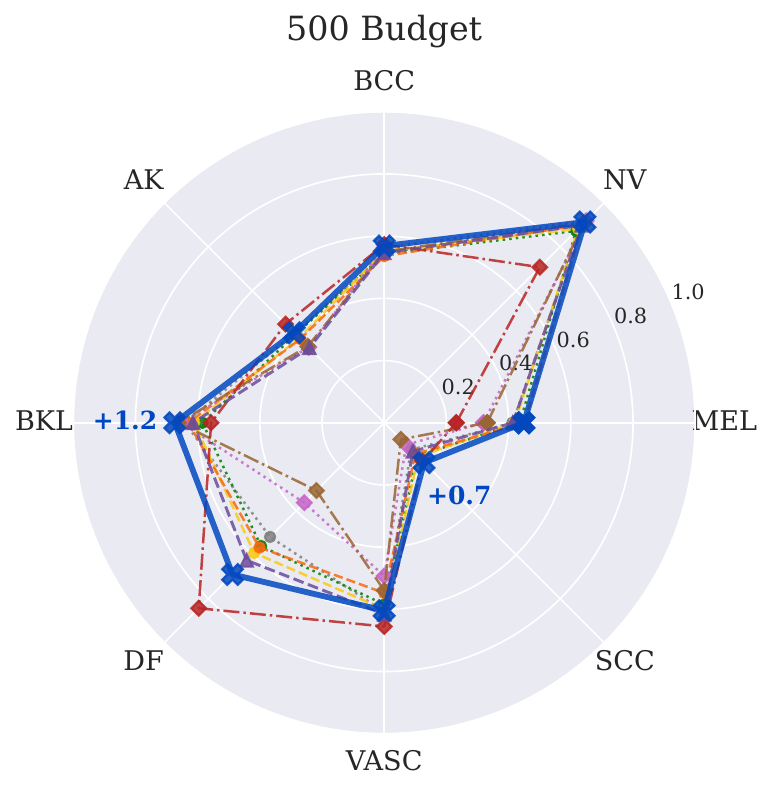} \\
    \end{tabular}
    \caption{\textbf{Per-class recall on FedISIC} in the OOD setting at four budgets $B\in\{50,150,300,500\}$ ($E{=}100$, 4 clinics), all at AL round $R{=}3$. The axes represent the 8 ID skin-lesion classes; blue annotations show the per-class gap (pp) between \texttt{FIDAL} and the best non-FIDAL baseline, averaged over 3 seeds.}
    \label{fig:fedisic-rareclasses}
\end{figure}

\begin{figure}[!t]
    \centering
    \includegraphics[width=\linewidth]{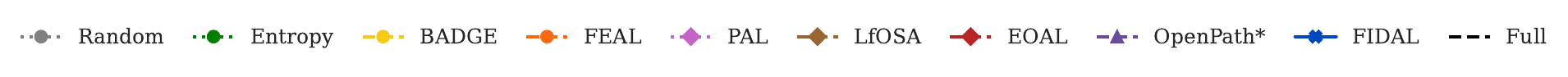}\\[-0.2em]
    \includegraphics[width=0.48\linewidth]{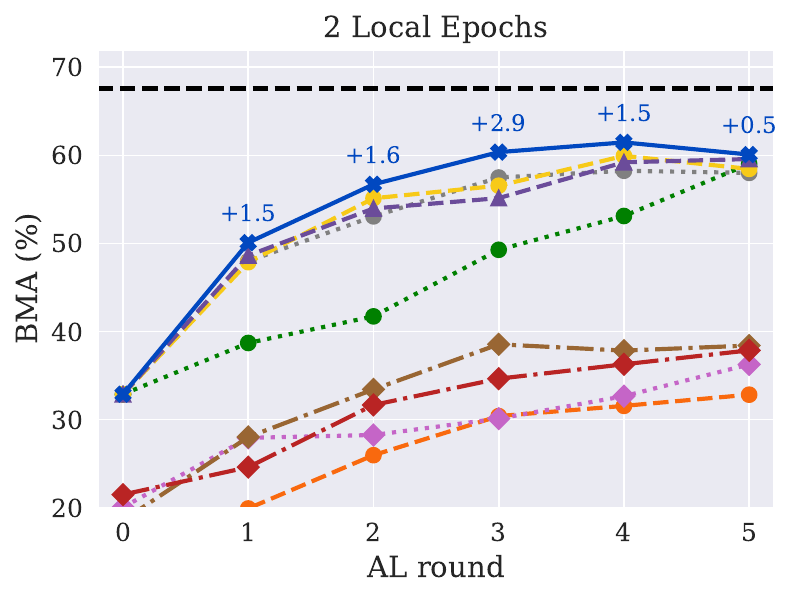}
    \includegraphics[width=0.48\linewidth]{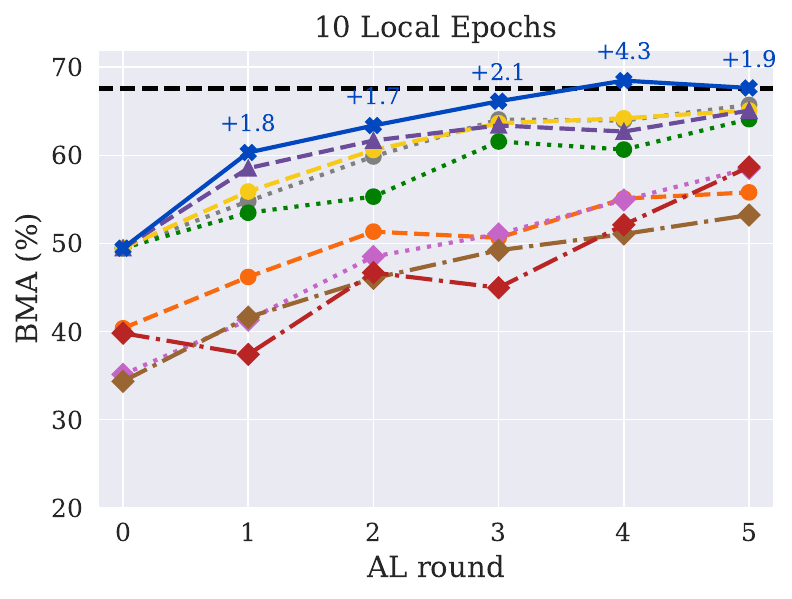}\\[0.2em]
    \includegraphics[width=0.48\linewidth]{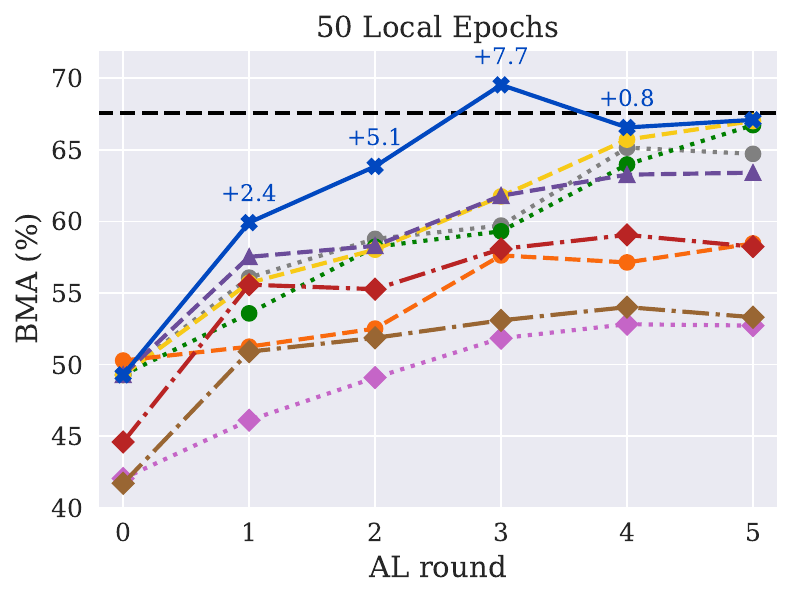}
    \includegraphics[width=0.48\linewidth]{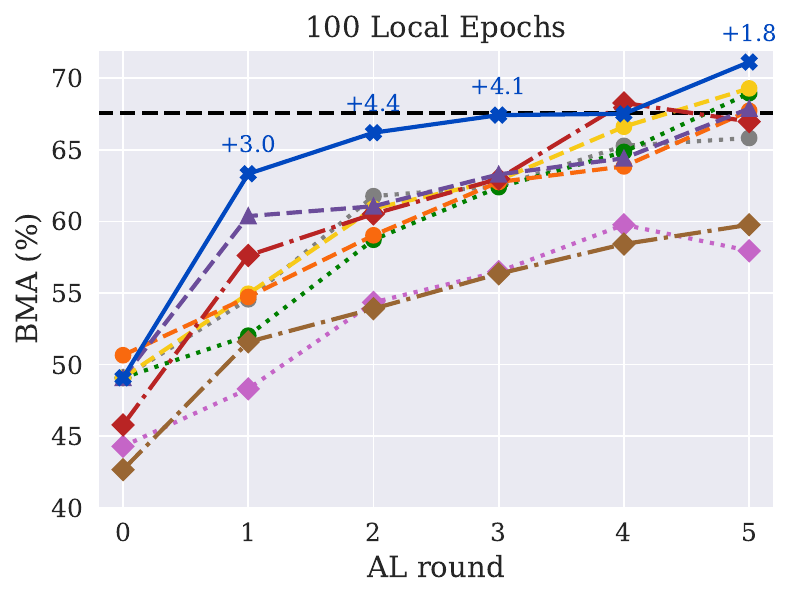}
    \caption{\textbf{Mean BMA trajectories across AL rounds on FedISIC OOD} for \texttt{FIDAL} and the eight baselines under local-epoch variation $E\in\{2,10,50,100\}$ at fixed $B{=}500$. Solid lines show the mean across 3 seeds, and blue annotations show the per-round gap between \texttt{FIDAL} and the best non-FIDAL baseline when the gap exceeds $0.1$\,pp. The dashed black line marks the fully supervised BMA reference ($\approx 67.71\%$). Budget-variation results at $E{=}100$ are reported in Supplementary Table~S3.}
    \label{fig:fedisic-budget-epoch-discussion}
\end{figure}

\textbf{Expert time.} The lower OOD query rate accumulates across rounds. On FedEMBED, if each avoided query had taken the mean mammogram reading time reported by \citet{partridge2024long} ($\approx 59$\,s), the avoided OOD queries correspond to roughly $7$--$29$ hours of expert reading, scaling linearly with cohort size and budget. For dermoscopy and histology patches we report only avoided counts, because we are not aware of a validated per-image reading time that maps onto our query units.

\textbf{Rare classes first.} At an equal budget, purity becomes label yield: \texttt{FIDAL} acquires $\approx 1.7\times$ more ID labels than the uncertainty baselines on FedISIC ($81.9\%$ versus $\leq 49\%$ purity) and $+10$\,pp more on FedEMBED ($92.3\%$ versus $82.0\%$ for \texttt{Entropy}). The extra labels go where they matter most. The per-class recall radars (Fig.~\ref{fig:fedisic-rareclasses}) show a ``rare-first'' progression: at $B{=}50$, \texttt{FIDAL} leads the best baseline by $+20.7$\,pp on DF ($1.0\%$ prevalence) and $+14.4$\,pp on VASC ($1.2\%$), paying with the majority NV class ($-4.8$\,pp); at $B{=}150$ the gains reach the middle tier (DF $+10.6$, AK $+9.6$, SCC $+5.0$\,pp) and at $B{=}300$ AK ($+9.9$) and BKL ($+4.2$\,pp); by $B{=}500$ the baselines have recovered the scarce classes and the advantage consolidates into overall balanced accuracy (NV matched at $-0.6$\,pp). This is the same effect the ablation isolates: the diversity term is the component whose removal costs the most accuracy.

\textbf{Few local epochs.} The largest gaps appear when the local models are least trained (Fig.~\ref{fig:fedisic-budget-epoch-discussion}). At $E{=}2$, \texttt{Random}, \texttt{Entropy}, \texttt{BADGE}, and \texttt{OpenPath*} stay in the $58$--$60\%$ band, \texttt{FEAL} falls to $32.83\%$, and the detector-based methods to $36$--$39\%$, while \texttt{FIDAL} reaches $61.47\%$, $1.5$\,pp above \texttt{BADGE} and $22.9$--$28.6$\,pp above the evidential and detector-based methods. One plausible reason is the source of the selection signal: \texttt{FEAL}'s evidential uncertainty and the auxiliary detectors of \texttt{PAL}, \texttt{LfOSA}, and \texttt{EOAL} depend on the learned local representation, whereas the frozen FM-coverage gate does not, although the experiment does not isolate this mechanism. Full fine-tuning (EfficientNet-B0, DenseNet-121) is also more epoch-sensitive than the frozen-encoder linear probe of FedEMBED, so freezing the encoder is one way to reduce epoch sensitivity under limited on-device compute; when the backbone must adapt, the gate keeps a selection signal that does not depend on it.

\subsection{Configuring the gate in practice}\label{subsec:hyperparam}
\begin{figure*}[!t]
  \centering
  \begin{tabular}{@{}cc@{}}
    \textbf{FedISIC} & \textbf{FedEMBED}\\
    \includegraphics[width=0.48\linewidth]{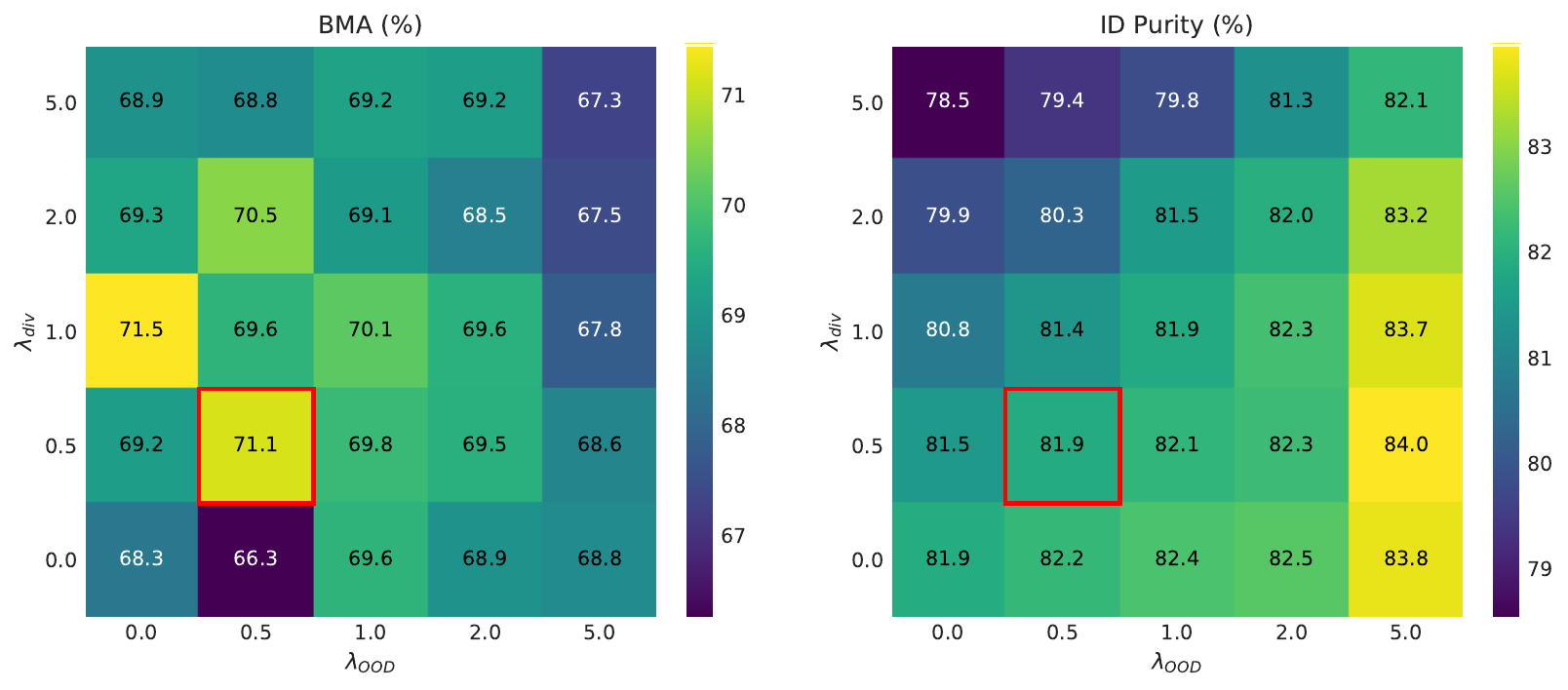} &
    \includegraphics[width=0.48\linewidth]{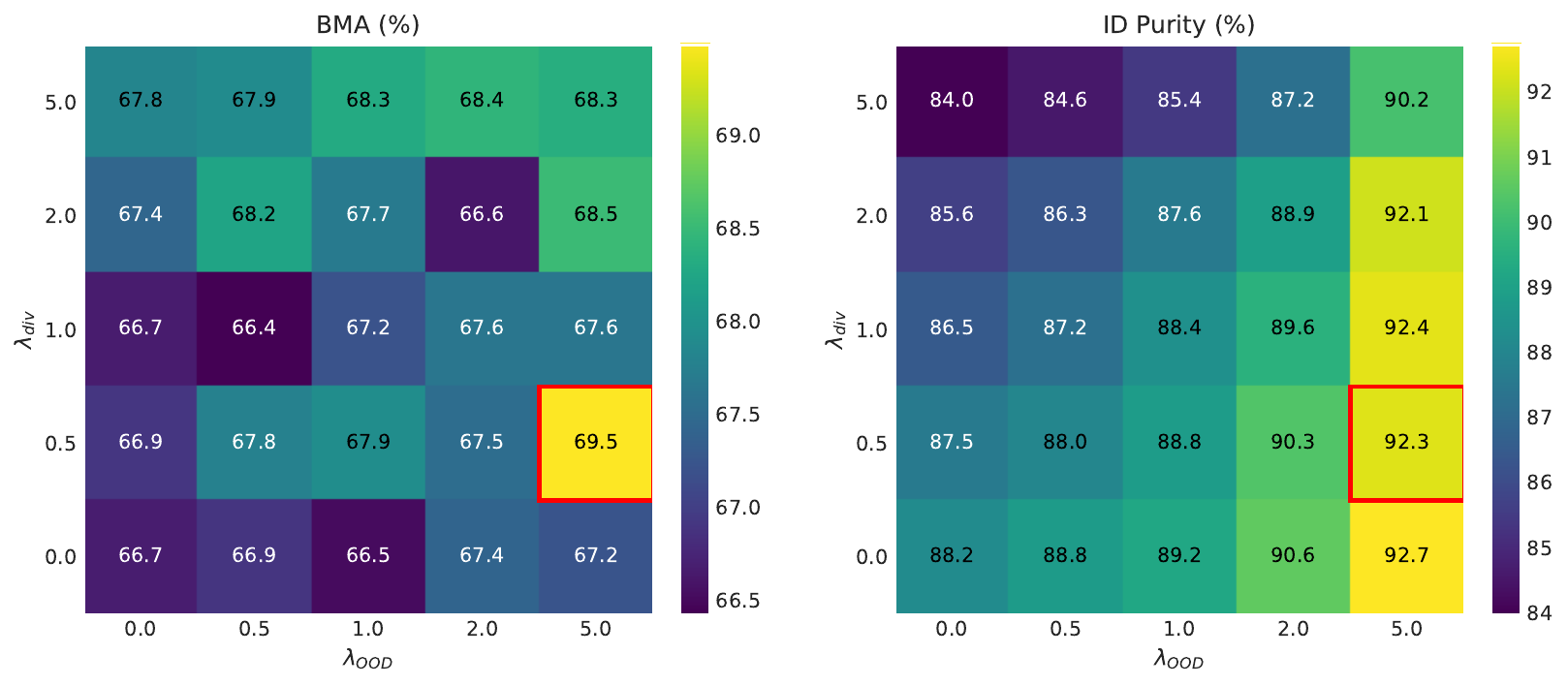} \\
    \includegraphics[width=0.48\linewidth]{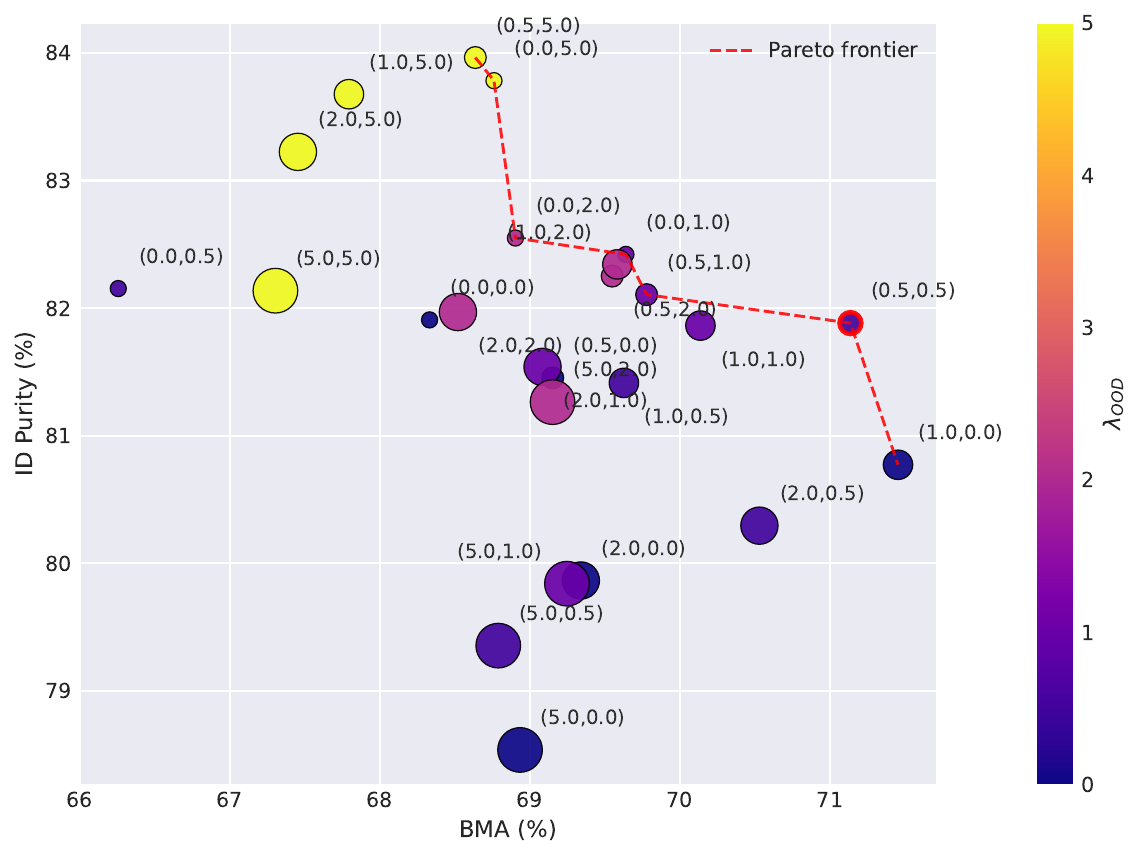} &
    \includegraphics[width=0.48\linewidth]{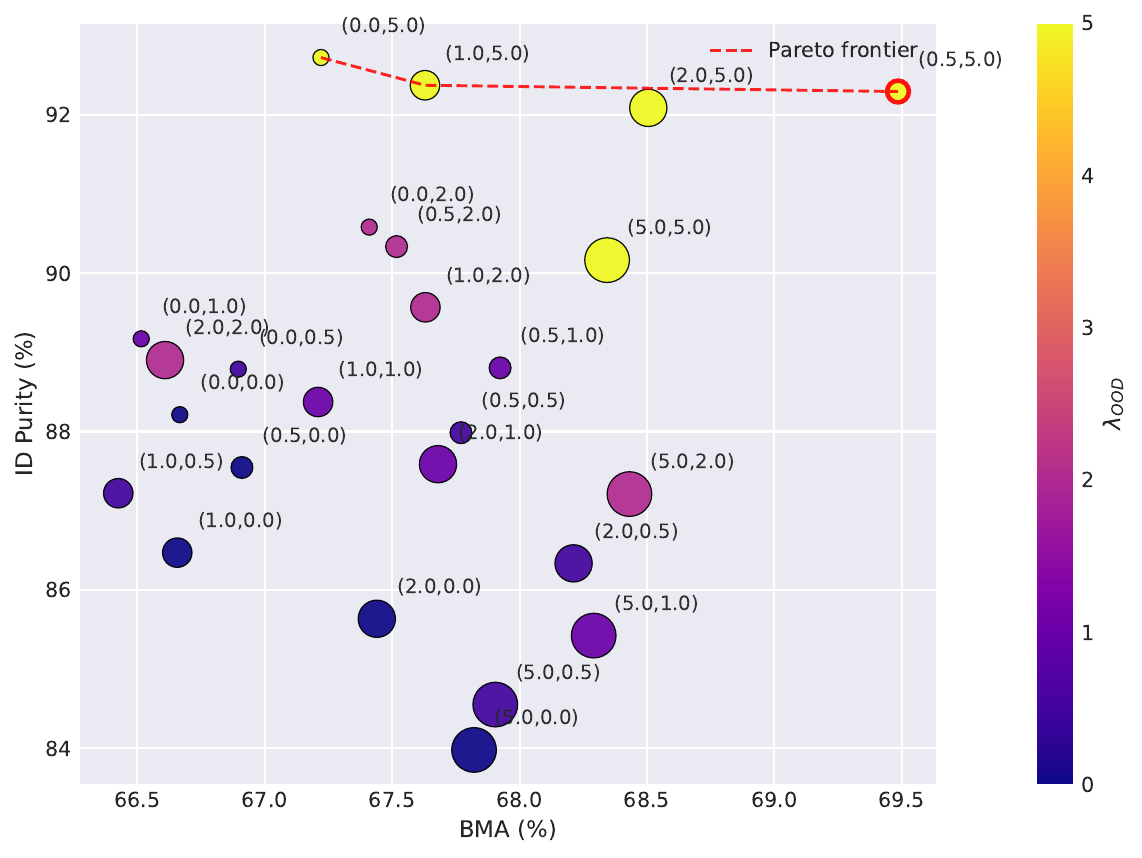} \\
  \end{tabular}
  \caption{\textbf{\texttt{FIDAL} sensitivity to $(\lambda_{\text{div}},\lambda_{\text{OOD}})$ on a $5{\times}5$ grid.} \textbf{Top}: heatmaps of BMA (left) and ID purity (right); the red box marks the adopted headline cell. \textbf{Bottom}: BMA-versus-ID purity Pareto scatter (color = $\lambda_{\text{OOD}}$, size = $\lambda_{\text{div}}$; red dashed line = frontier). \textbf{(Left)} FedISIC at $R{=}5$; \textbf{(Right)} FedEMBED at $R{=}7$. The selected headline cells are FedISIC $(0.5,0.5)$ (all components active, smallest seed spread) and FedEMBED $(0.5,5.0)$ (on the ID purity Pareto frontier, with headline BMA $0.05$\,pp below the supervised reference).}
  \label{fig:sensitivity-recovered}
\end{figure*}

FedEMBED illustrates the regime that controlled benchmarks rarely capture: each scanner carries a different artifact mix (prevalence $4.83$--$13.12\%$, Supplementary Table~S2), so a gate keyed to a fixed OOD vocabulary or a global threshold may not transfer across institutions. \texttt{FIDAL}'s gate is refit per client and per round on the local coverage distribution, with no shared prompts, exemplars, or pool-wide statistics; on FedISIC the same procedure rejects the DDI/Fitzpatrick17k clinical photographs without a manual prompt. Whether the gate suffices is governed by FM--OOD separability (Fig.~\ref{fig:fm_story}): where the encoder places the OOD off the ID manifold, the Otsu gate alone filtered strongly in our settings; where it does not, the explicit $S_{\text{OOD}}$ penalty must carry the purity, which is why FedEMBED uses a larger $\lambda_{\text{OOD}}$.

The $5{\times}5$ grid over $(\lambda_{\text{div}},\lambda_{\text{OOD}})\in\{0,0.5,1,2,5\}^{2}$ on FedISIC and FedEMBED (Fig.~\ref{fig:sensitivity-recovered}) makes this concrete. On FedISIC the best BMA, $(71.45 \pm 0.88)\%$, occurs at $(1.0,0.0)$, consistent with the gate providing most of the rejection signal; the all-components cell $(0.5,0.5)$ reaches $(71.14 \pm 0.54)\%$ with the smallest seed spread and is used as headline. On FedEMBED, BMA varies by at most $1.2$\,pp across the upper-$\lambda$ cells at $R{=}7$, so we select by ID purity, which peaks along the $\lambda_{\text{OOD}}{=}5$ column; the adopted $(0.5,5.0)$ lies on the purity Pareto frontier. FedCamelyon was not swept and uses the symmetric default $(1.0,1.0)$. As a pilot heuristic for a new dataset: run the one-round gate comparison of Fig.~\ref{fig:fm_story}; keep $\lambda_{\text{div}}\in[0.5,1.0]$, which retains at least $\approx$$95\%$ of the BMA optimum on both swept datasets; and choose $\lambda_{\text{OOD}}\in[0,1]$ if the gate catches $\gtrsim 90\%$ of the OOD at $\leq 2\%$ ID loss, or $[2,5]$ if it does not. These ranges describe the evaluated datasets, not universal cutoffs.

\subsection{Qualitative assessment of what FIDAL selects and rejects}\label{subsec:qualitative}
\begin{figure*}[!t]
    \centering
    \begin{minipage}{0.03\linewidth} \hfill \end{minipage}
    \begin{minipage}{0.185\linewidth} \centering \small \textbf{Easy ID} \\ \scriptsize (Less informative) \end{minipage}
    \begin{minipage}{0.185\linewidth} \centering \small \textbf{Hard OOD} \\ \scriptsize (queried OOD) \end{minipage}
    \begin{minipage}{0.185\linewidth} \centering \small \textbf{Diverse ID} \\ \scriptsize (Informative) \end{minipage}
    \begin{minipage}{0.185\linewidth} \centering \small \textbf{Filtered ID} \\ \scriptsize (ground-truth ID) \end{minipage}
    \begin{minipage}{0.185\linewidth} \centering \small \textbf{Filtered OOD} \\ \scriptsize (ground-truth OOD) \end{minipage}
    \vspace{0.1cm}

    \begin{minipage}{0.03\linewidth}\rotatebox{90}{\textbf{FedISIC}}\end{minipage}
    \begin{minipage}{0.185\linewidth}
        \setlength{\fboxsep}{2pt}
        \fcolorbox{idblue}{white}{\includegraphics[width=\dimexpr\linewidth-4.8pt\relax]{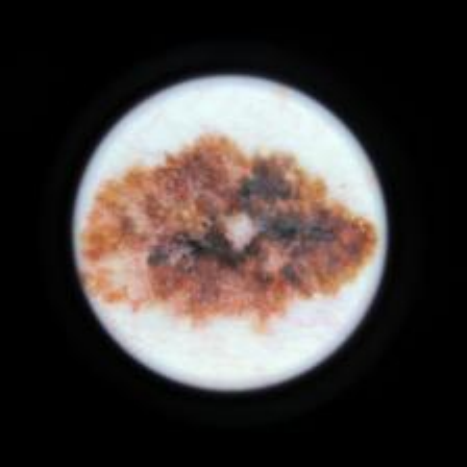}}
        \centerline{\scriptsize $S_{\text{ID}}{=}0.981$}
    \end{minipage}
    \begin{minipage}{0.185\linewidth}
        \setlength{\fboxsep}{2pt}
        \fcolorbox{oodred}{white}{\includegraphics[width=\dimexpr\linewidth-4.8pt\relax]{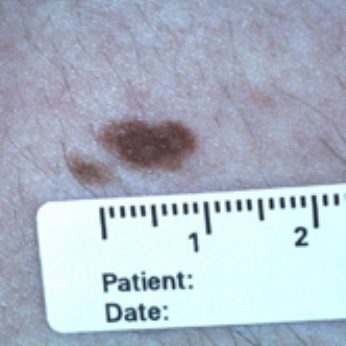}}
        \centerline{\scriptsize $S_{\text{OOD}}{=}0.975$}
    \end{minipage}
    \begin{minipage}{0.185\linewidth}
        \setlength{\fboxsep}{2pt}
        \fcolorbox{idblue}{white}{\includegraphics[width=\dimexpr\linewidth-4.8pt\relax]{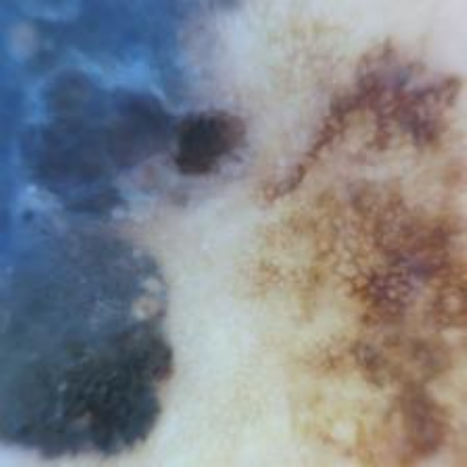}}
        \centerline{\scriptsize $S_{\text{ID}}{=}0.347$}
    \end{minipage}
    \begin{minipage}{0.185\linewidth}
        \setlength{\fboxsep}{2pt}
        \fcolorbox{idblue}{white}{\includegraphics[width=\dimexpr\linewidth-4.8pt\relax]{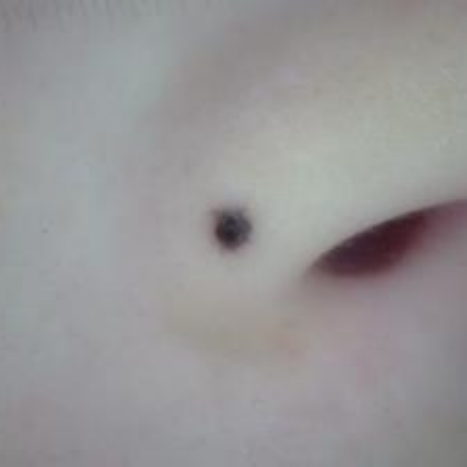}}
        \centerline{\scriptsize $G(\mathbf{x}){=}0.488$}
    \end{minipage}
    \begin{minipage}{0.185\linewidth}
        \setlength{\fboxsep}{2pt}
        \fcolorbox{green}{white}{\includegraphics[width=\dimexpr\linewidth-4.8pt\relax]{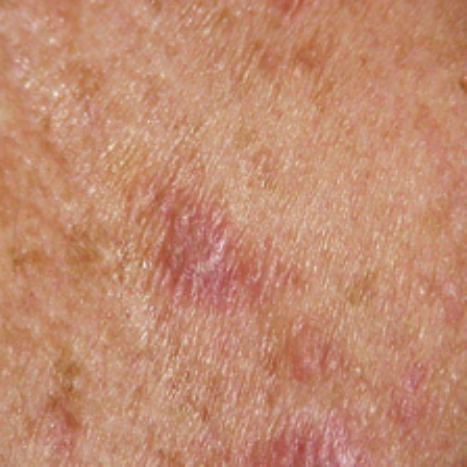}}
        \centerline{\scriptsize $G(\mathbf{x}){=}0.676$}
    \end{minipage}
    \vspace{0.2cm}

    \begin{minipage}{0.03\linewidth}\rotatebox{90}{\textbf{FedCamelyon}}\end{minipage}
    \begin{minipage}{0.185\linewidth}
        \setlength{\fboxsep}{2pt}
        \fcolorbox{idblue}{white}{\includegraphics[width=\dimexpr\linewidth-4.8pt\relax]{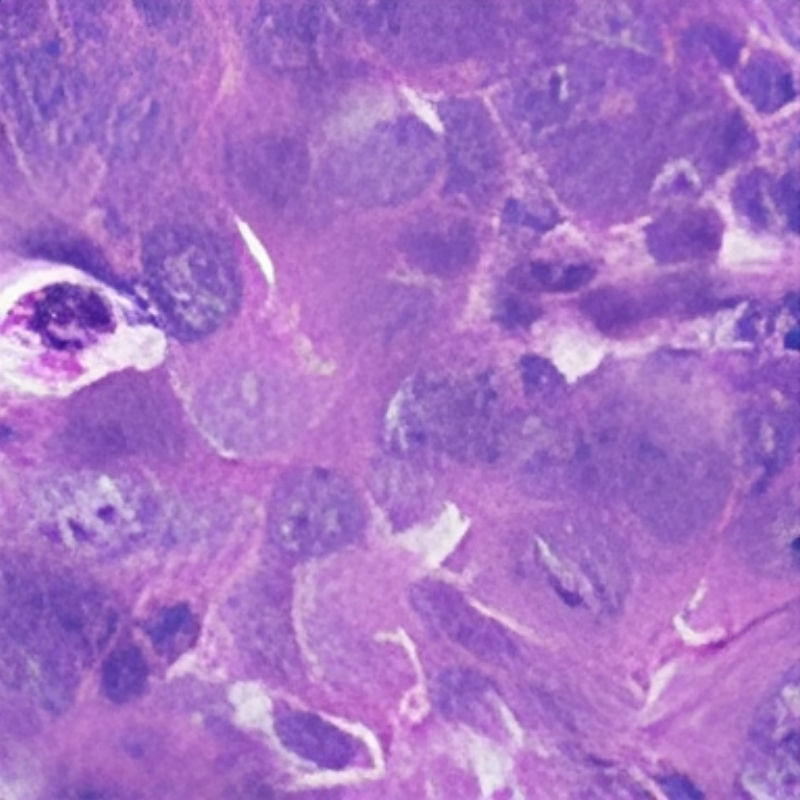}}
        \centerline{\scriptsize $S_{\text{ID}}{=}0.992$}
    \end{minipage}
    \begin{minipage}{0.185\linewidth}
        \setlength{\fboxsep}{2pt}
        \fcolorbox{oodred}{white}{\includegraphics[width=\dimexpr\linewidth-4.8pt\relax]{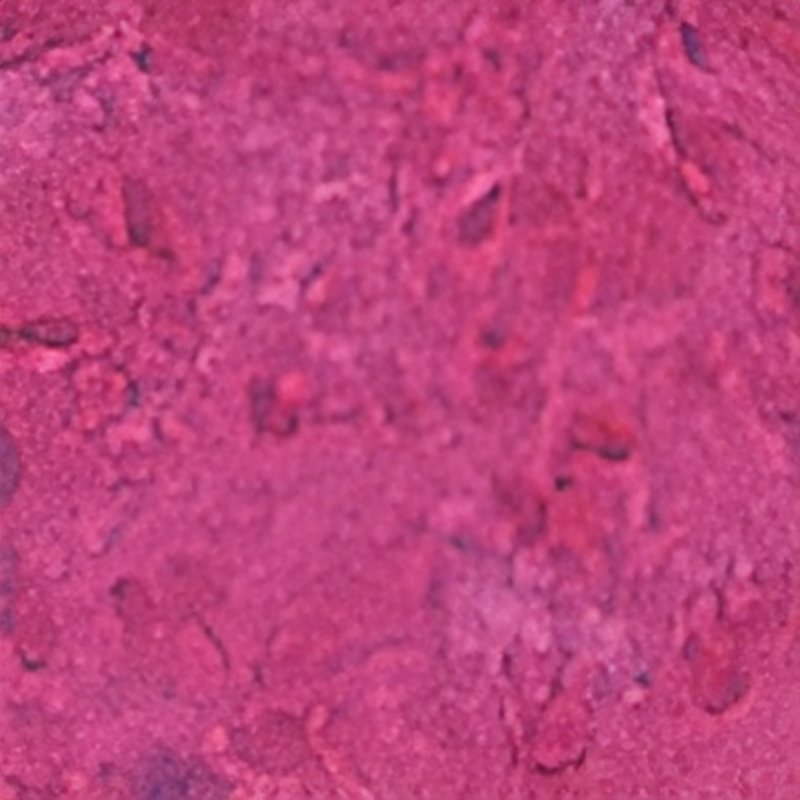}}
        \centerline{\scriptsize $S_{\text{OOD}}{=}0.991$}
    \end{minipage}
    \begin{minipage}{0.185\linewidth}
        \setlength{\fboxsep}{2pt}
        \fcolorbox{idblue}{white}{\includegraphics[width=\dimexpr\linewidth-4.8pt\relax]{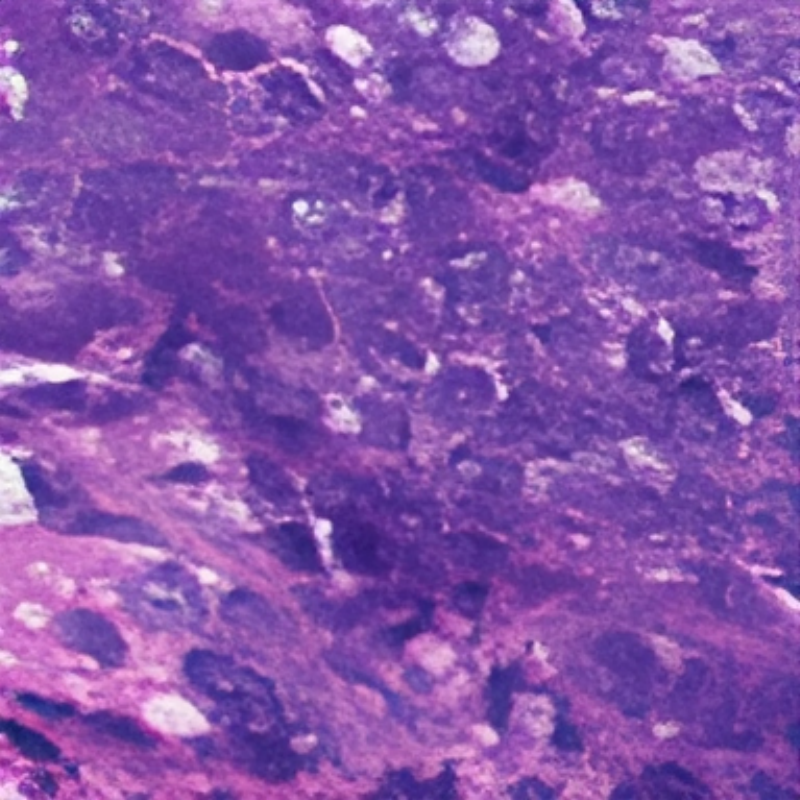}}
        \centerline{\scriptsize $S_{\text{ID}}{=}0.777$}
    \end{minipage}
    \begin{minipage}{0.185\linewidth}
        \setlength{\fboxsep}{2pt}
        \fcolorbox{idblue}{white}{\includegraphics[width=\dimexpr\linewidth-4.8pt\relax]{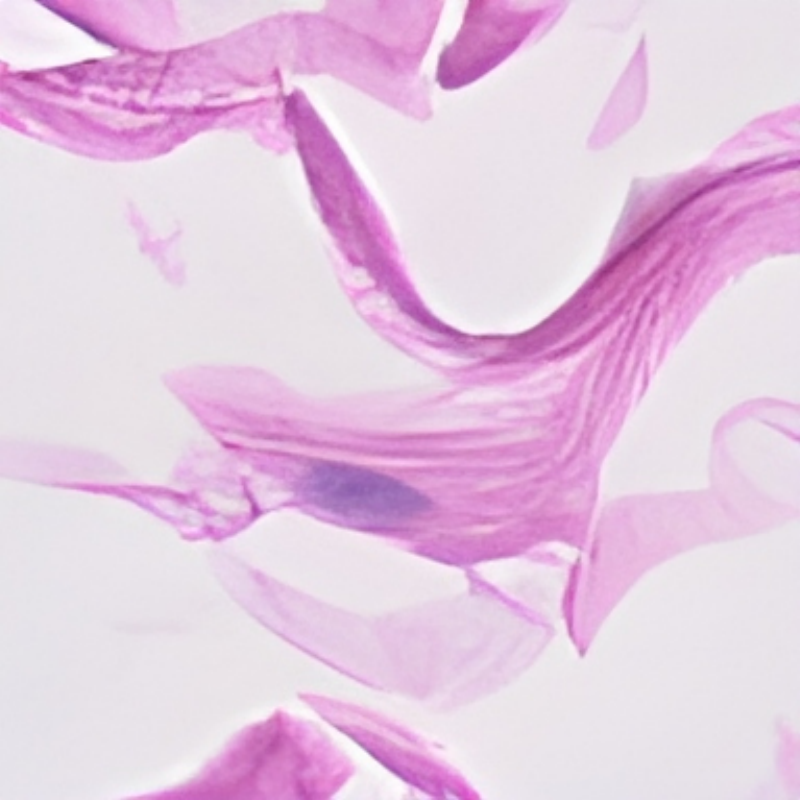}}
        \centerline{\scriptsize $G(\mathbf{x}){=}0.826$}
    \end{minipage}
    \begin{minipage}{0.185\linewidth}
        \setlength{\fboxsep}{2pt}
        \fcolorbox{green}{white}{\includegraphics[width=\dimexpr\linewidth-4.8pt\relax]{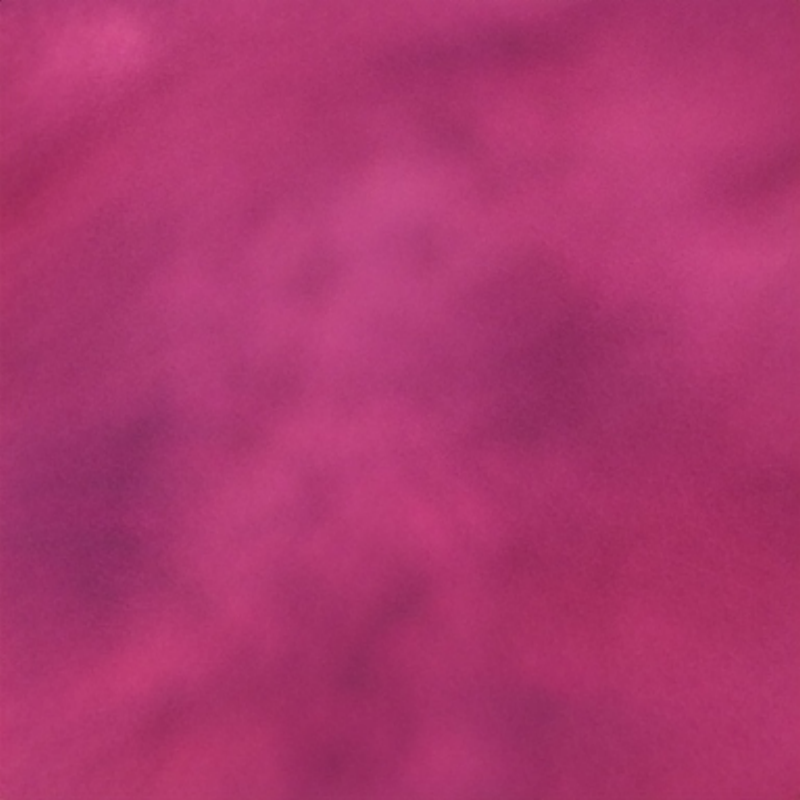}}
        \centerline{\scriptsize $G(\mathbf{x}){=}0.033$}
    \end{minipage}
    \vspace{0.2cm}

    \begin{minipage}{0.03\linewidth}\rotatebox{90}{\textbf{FedEMBED}}\end{minipage}
    \begin{minipage}{0.185\linewidth}
        \setlength{\fboxsep}{2pt}
        \fcolorbox{idblue}{white}{\includegraphics[width=\dimexpr\linewidth-4.8pt\relax]{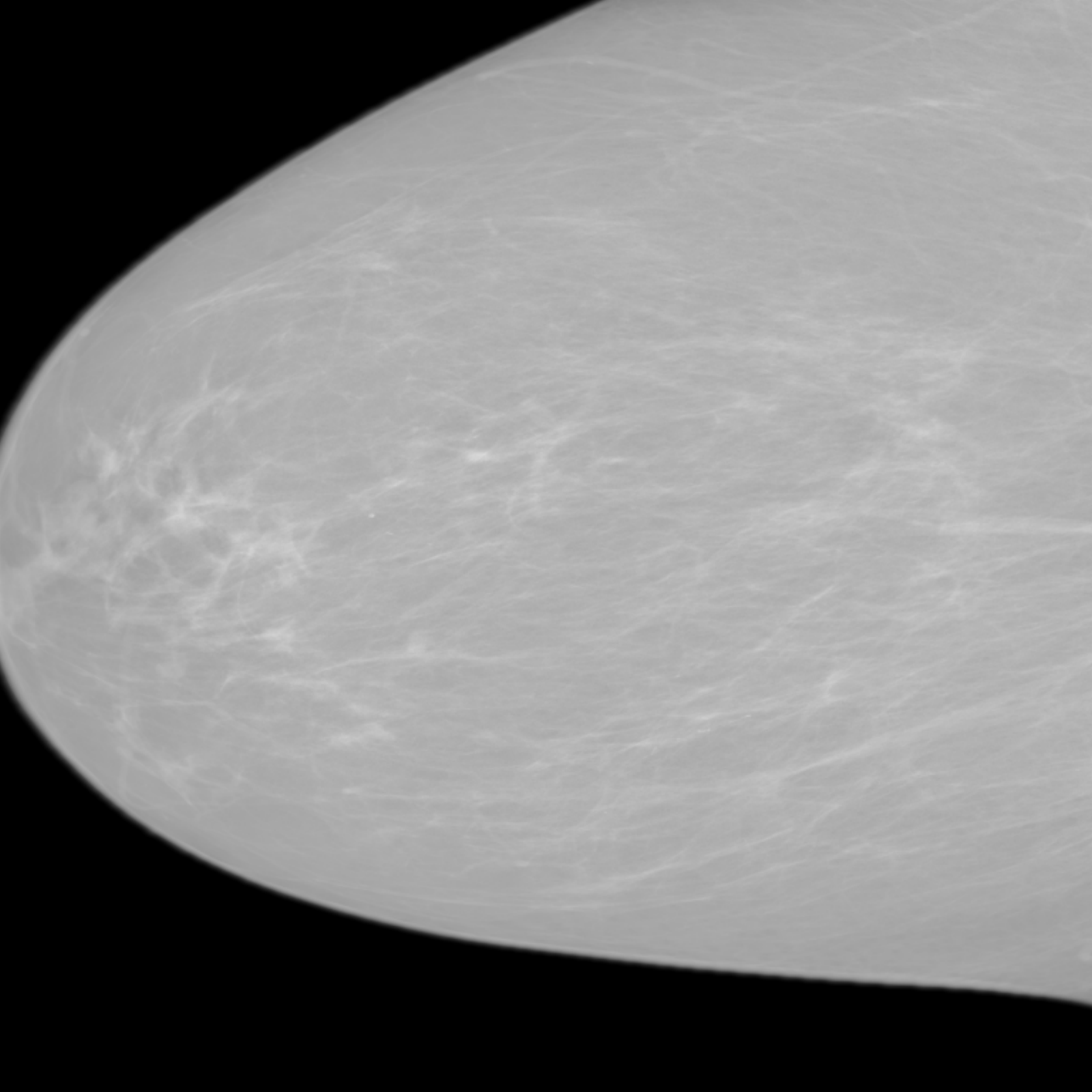}}
        \centerline{\scriptsize $S_{\text{ID}}{=}0.976$}
    \end{minipage}
    \begin{minipage}{0.185\linewidth}
        \setlength{\fboxsep}{2pt}
        \fcolorbox{oodred}{white}{\includegraphics[width=\dimexpr\linewidth-4.8pt\relax]{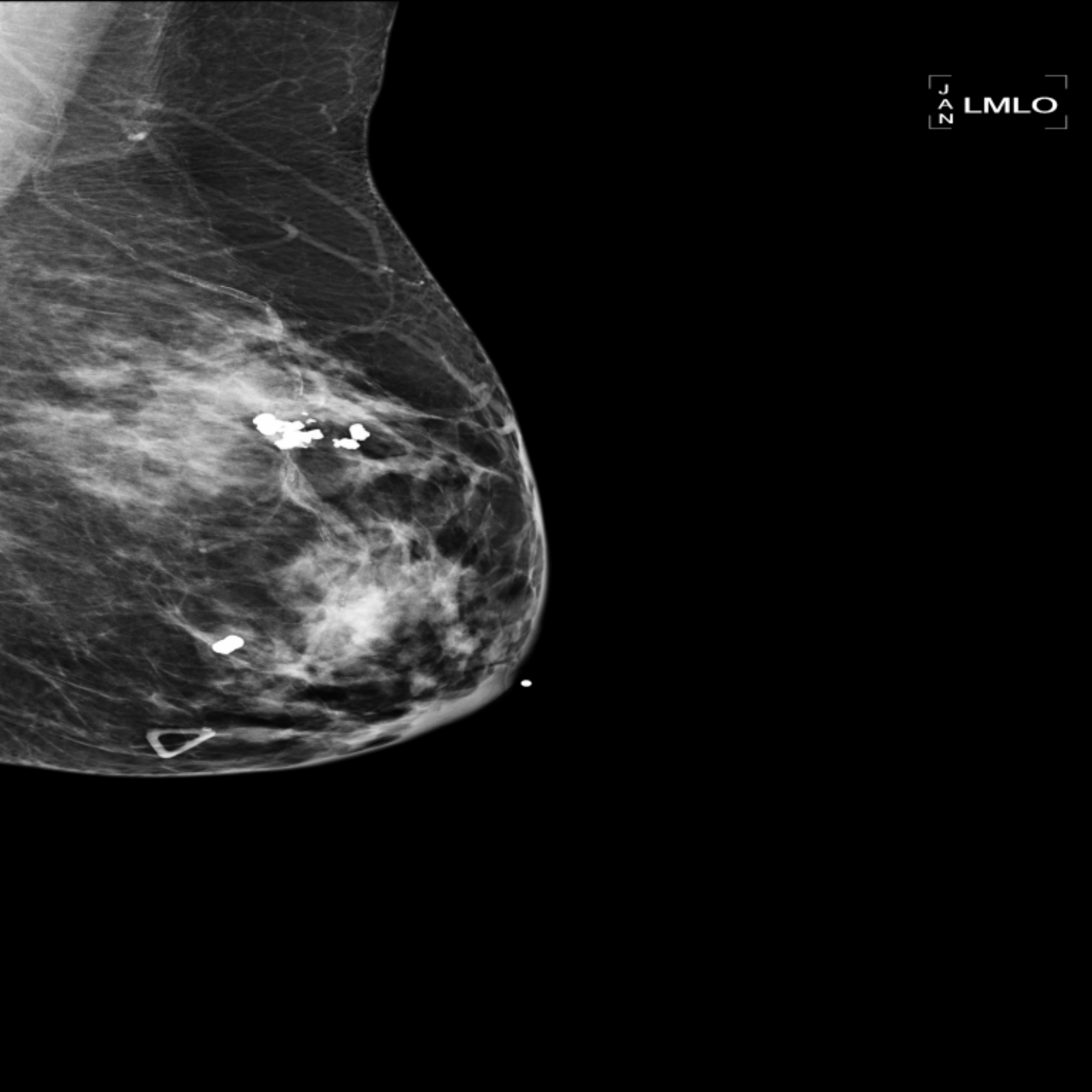}}
        \centerline{\scriptsize $S_{\text{OOD}}{=}0.918$}
    \end{minipage}
    \begin{minipage}{0.185\linewidth}
        \setlength{\fboxsep}{2pt}
        \fcolorbox{idblue}{white}{\includegraphics[width=\dimexpr\linewidth-4.8pt\relax]{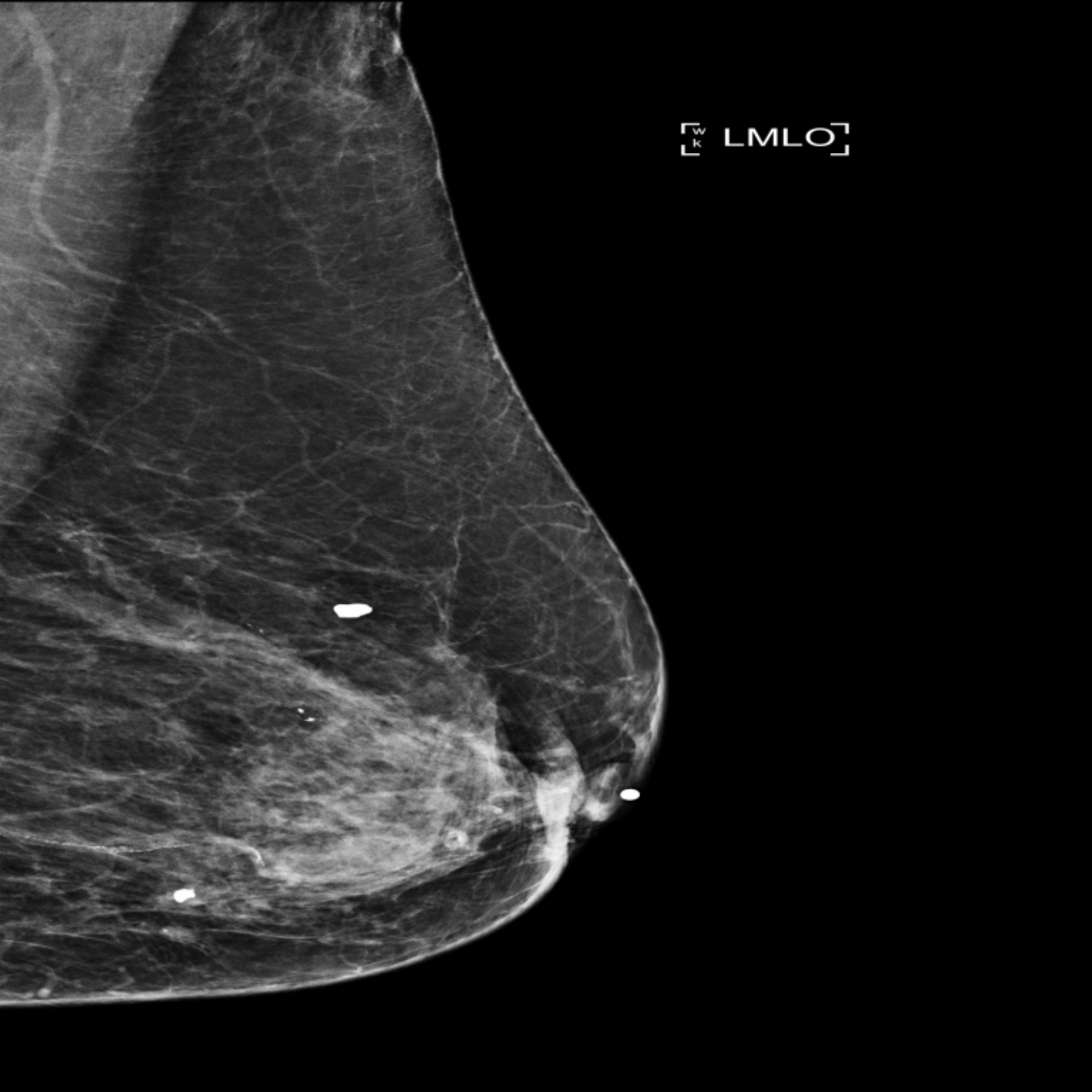}}
        \centerline{\scriptsize $S_{\text{ID}}{=}0.425$}
    \end{minipage}
    \begin{minipage}{0.185\linewidth}
        \setlength{\fboxsep}{2pt}
        \fcolorbox{idblue}{white}{\includegraphics[width=\dimexpr\linewidth-4.8pt\relax]{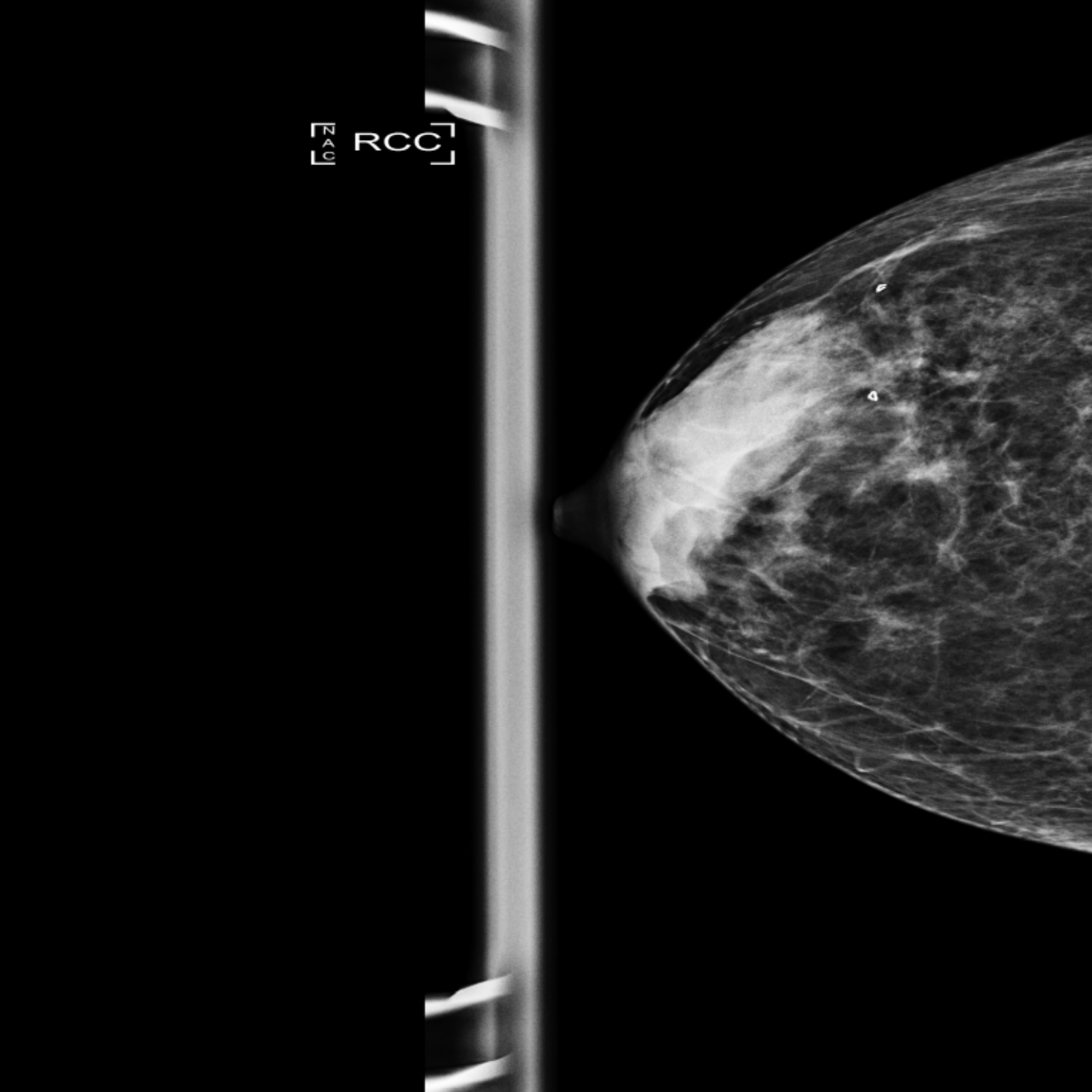}}
        \centerline{\scriptsize $G(\mathbf{x}){=}0.823$}
    \end{minipage}
    \begin{minipage}{0.185\linewidth}
        \setlength{\fboxsep}{2pt}
        \fcolorbox{green}{white}{\includegraphics[width=\dimexpr\linewidth-4.8pt\relax]{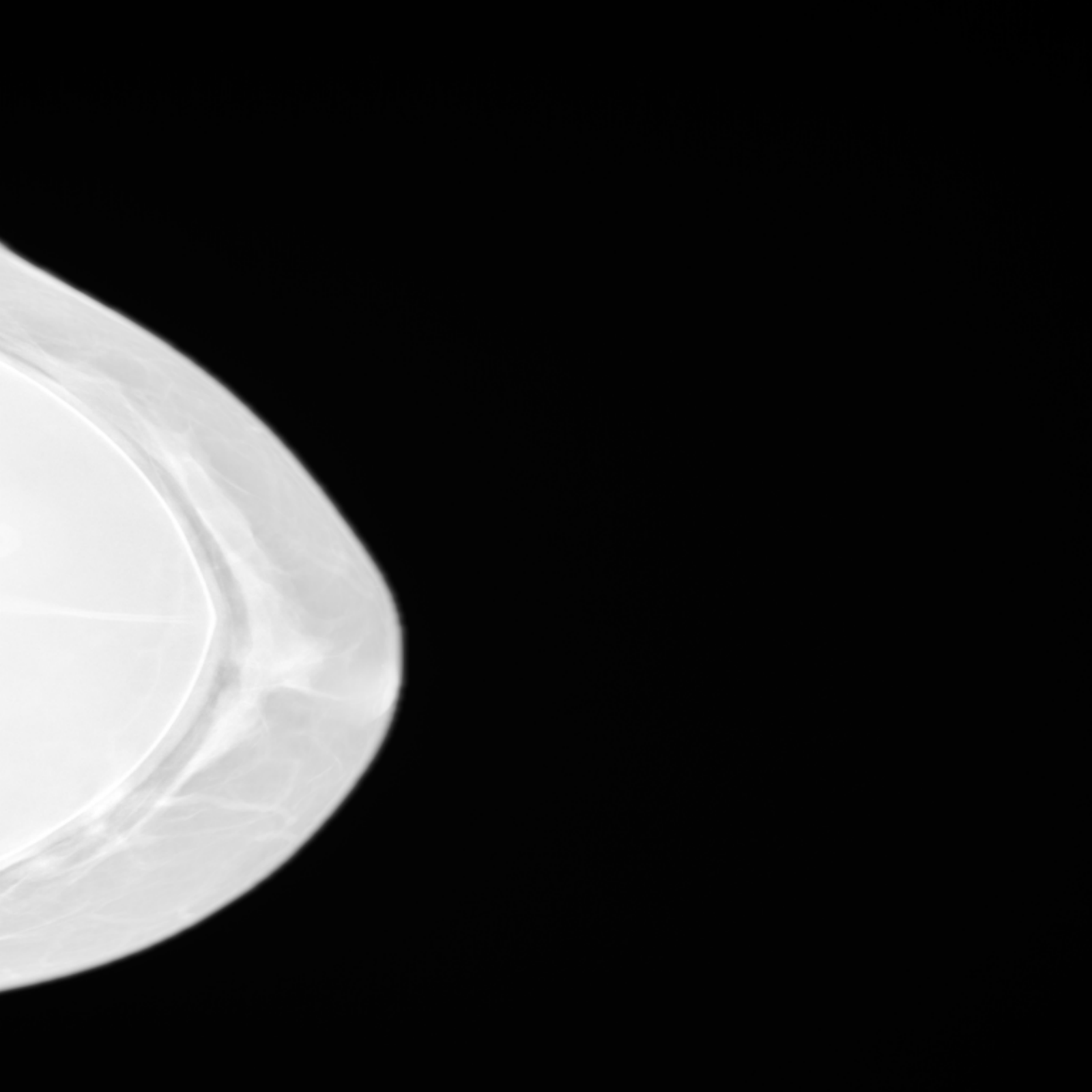}}
        \centerline{\scriptsize $G(\mathbf{x}){=}0.276$}
    \end{minipage}
    \vspace{0.2cm}

    \caption{\textbf{Qualitative exemplars selected or gated by \texttt{FIDAL} at the final AL round.} Rows correspond to datasets. The first three columns show Easy ID, Hard OOD, and Diverse ID samples selected by \texttt{FIDAL}, together with their scores. The last two columns show ground-truth ID and OOD samples rejected before querying by the FM-coverage Otsu gate, together with the coverage signal $G(\mathbf{x})\!\in\![0,1]$. Blue frames denote ground-truth ID samples, red frames denote OOD samples that passed the gate and were queried, and green frames denote OOD samples correctly rejected by the gate.}
    \label{fig:qualitative_analysis_Rfinal}
\end{figure*}

Fig.~\ref{fig:qualitative_analysis_Rfinal} shows what the scores mean in images. The ``Easy ID'' picks are canonical cases (a centered dermoscopy nevus, $S_{\text{ID}}\!\approx\!0.981$), the ``Diverse ID'' picks atypical ones (a multicolored lesion, $S_{\text{ID}}\!\approx\!0.347$; a mammogram with granular microcalcifications, $S_{\text{ID}}\!\approx\!0.425$), and the ``Hard OOD'' samples that pass the gate closely resemble labeled artifacts (a clinical record with a paper ruler, $S_{\text{OOD}}\!\approx\!0.975$; a blood-dominated patch, $0.991$; a breast implant, $0.918$, the EMBED artifact class with the largest reported performance decrease~\citep{10981259}). The gate correctly rejects low-coverage OOD such as a near-uniform blur tile ($G(\mathbf{x})\!\approx\!0.033$) and an almost empty implant acquisition ($0.276$), but also filters some true-ID cases near the coverage boundary, such as a nevus on facial skin ($0.488$) or a ruptured-membrane fragment ($0.826$). The gate is therefore conservative, trading a little ID recall for the exclusion of low-coverage OOD; the $S_{\text{OOD}}$ penalty then handles the OOD that passes, without another hard threshold.

\subsection{Scope and limitations}\label{subsec:limitations}
\textbf{Where FIDAL does not lead.} On the ID purity axis alone, \texttt{LfOSA} attains an equal or higher score (FedCamelyon $90.7\%$ versus $83.8\%$; FedEMBED $92.6\%$ versus $92.3\%$) and at $R{=}10$ acquires more ID samples than \texttt{FIDAL} ($24{,}933$ versus $23{,}035$ on FedCamelyon; $19{,}780$ versus $19{,}515$ on FedEMBED), but with lower classification performance ($-2.93$ and $-10.00$\,pp, respectively); our experiments do not isolate why. In the scarce-label regime ($B{=}50$), \texttt{FEAL} leads on FedCamelyon ACC ($(92.20 \pm 0.16)\%$ versus $(91.82 \pm 0.45)\%$) and \texttt{OpenPath*} on FedEMBED BMA ($(64.92 \pm 1.24)\%$ versus $(60.90 \pm 1.59)\%$), while \texttt{FIDAL} retains the highest ID purity ($96.1\%$ and $95.5\%$). We therefore report the joint accuracy--ID purity frontier (Fig.~\ref{fig:pareto_bma_idp_3datasets_global}) rather than either axis alone, and recommend a small per-dataset pilot for $(\lambda_{\text{div}},\lambda_{\text{OOD}})$ rather than a fixed default.

\textbf{Privacy and regulatory compliance.} Similarity scoring and evidential uncertainty (Eq.~\eqref{eq:fidal_uncertainty_calib}) are computed within each client; feature vectors, embeddings, and raw data are never shared with the server or other clients. This supports data minimization but does not by itself establish compliance with the General Data Protection Regulation (GDPR) or HIPAA, which depends on deployment infrastructure, institutional policies, and jurisdiction; integration with secure aggregation~\citep{10.1145/3133956.3133982} was not evaluated.

\textbf{Label noise and diagnostic complexity.} FedISIC and FedCamelyon assume a reliable labeling oracle. EMBED contains approximately $20\%$ ``ambiguous'' ROIs~\citep{jeong2023emory}; this does not affect the breast-density labels used here, but the aleatoric term (Eq.~\eqref{eq:aleatoric_feal_calib}) does not explicitly model label noise, and we do not model settings in which the ID/OOD definition itself drifts across institutions or guidelines, which calls for longitudinal, multi-annotator OS-FAL benchmarks.

\textbf{Sex- and gender-based analysis.} Sex and gender were not used as variables in this study: the federated partitions are defined by acquisition site or scanner and the OOD categories in acquisition terms, so we did not stratify accuracy, ID purity, or gate rejection rates by patient sex or gender, which limits the generalizability of our findings. Deployment-oriented follow-up work should analyze acquisition patterns and model performance by sex, gender, skin tone, and age; the FedISIC OOD sources~\citep{daneshjou2022disparities,groh2021evaluating} provide variation in imaging conditions and skin tones that we did not use for subgroup-stratified evaluation.

\section{Conclusion}\label{sec:conclusion}
\FloatBarrier
We presented FIDAL, an OS-FAL framework for medical imaging pools that contain irrelevant or OOD samples. The method separates two decisions that are often conflated, whether a local sample is suitable for annotation and, if so, how informative it is, by combining an adaptive, client-specific gate derived from FM coverage with calibrated uncertainty, support-aware diversity, and similarity to previously identified OOD samples. Across FedISIC, FedCamelyon, and FedEMBED, this design produced a consistent balance between predictive performance and annotation efficiency, reaching $81.9$--$92.3\%$ ID purity and reducing OOD queries by factors of $1.3$--$3.8$ compared with accuracy-competitive baselines. Future work will extend FIDAL to more heterogeneous federated environments with diverse modalities, variable annotation budgets, and realistic label-noise patterns, including settings where the ID/OOD boundary itself drifts across institutions and over time.

\section*{CRediT authorship contribution statement}
\textbf{David Due\~nas Gaviria:} Conceptualization, Methodology, Software, Formal analysis, Visualization, Writing -- original draft.
\textbf{Shadi Albarqouni:} Conceptualization, Methodology, Supervision, Funding acquisition, Writing -- review and editing.

\section*{Data availability}
No new primary imaging data were generated in this study; all experiments use previously published, publicly available datasets, which are cited in the reference list. FedISIC is derived from ISIC-2019 under the standard FLamby federated partition~\citep{ogier2022flamby}, with OOD samples drawn from DDI~\citep{daneshjou2022disparities} and Fitzpatrick17k~\citep{groh2021evaluating}; FedCamelyon is derived from CAMELYON17~\citep{bandi2018detection} with OOD samples from HistoArtifacts~\citep{kanwal2024equipping}. Both are built entirely on open data, and we release the exact CSV split files together with our code. FedEMBED is derived from the EMBED dataset~\citep{jeong2023emory}, which is available to credentialed users under a standard data-use agreement; because these images are access-controlled, we cannot redistribute them or our split files, and we instead release the preprocessing and split-generation code that reproduces our partitions from the licensed data. Our full implementation, together with the FedISIC and FedCamelyon split files, will be made publicly available at \url{https://github.com/albarqounilab/fidal}.

\section*{Supplementary material}
Supplementary material (Sections~S1--S6) follows the reference list at the end of this document and contains the per-client OOD composition of FedCamelyon and FedEMBED (Tables~S1--S2), training details and the federated \texttt{OpenPath*} prompt protocol (Section~S2), per-client Pareto trajectories (Fig.~S1), per-round labeled-pool composition (Figs.~S2--S4), the full budget and local-epoch variation tables (Tables~S3--S5), and the protocol and caveats of the per-class recall analysis (Section~S6).

{
    \small
    \bibliographystyle{ieeenat_fullname}
    \bibliography{fidal_references}
}

\input{supplementary}

\end{document}

%% file: preamble.tex
\usepackage{amsfonts}
\usepackage{multirow}
\usepackage{placeins}

\graphicspath{{./figures/}}

\newcommand{\mathds}[1]{\mathbb{#1}}

\definecolor{warning}{HTML}{FF8C00}
\definecolor{green}{HTML}{2CB41F}
\definecolor{idblue}{HTML}{1F77B4}
\definecolor{oodred}{HTML}{D62728}

%% file: supplementary.tex
\clearpage
\onecolumn
\setcounter{section}{0}\setcounter{subsection}{0}\setcounter{figure}{0}\setcounter{table}{0}
\renewcommand{\thesection}{S\arabic{section}}
\renewcommand{\thefigure}{S\arabic{figure}}
\renewcommand{\thetable}{S\arabic{table}}
\setlength{\emergencystretch}{2em}

\begin{center}
{\Large\bfseries Supplementary Material}\\[4pt]
\end{center}
\vspace{0.5em}

\section{Open-set benchmark composition}\label{ssec:composition}

Tables~\ref{tab:fedcamelyon_ood} and~\ref{tab:fedembed_ood} give the per-client composition of the OOD pools summarized in main-text Section~4.1: the HistoArtifacts pool injected into each FedCamelyon center at $38.5\%$ prevalence, broken down by artifact type, and the organically occurring, expert-annotated artifacts of FedEMBED, whose prevalence and type mix differ by scanner.

\begin{table}[!htbp]
\centering
\caption{Composition of the injected OOD pool for \textbf{FedCamelyon} across the five centers. OOD is injected at $38.5\%$ of each center's pool.}
\label{tab:fedcamelyon_ood}
\resizebox{\linewidth}{!}{%
\begin{tabular}{l c cccc}
\toprule
\multirow{2}{*}{\textbf{Center}} &
\textbf{Total OOD} &
\multicolumn{4}{c}{\textbf{Artifact breakdown}} \\
\cmidrule(lr){3-6}
 & & \textbf{Blood} & \textbf{Blur} & \textbf{Bubbles} & \textbf{Tissue damage} \\
\midrule
client\,1 &  4{,}754 & 3{,}185 (25.8\%) &   591 (4.8\%) &   438 (3.5\%) &   540 (4.4\%) \\
client\,2 &  2{,}791 & 1{,}855 (25.6\%) &   360 (5.0\%) &   261 (3.6\%) &   315 (4.3\%) \\
client\,3 &  6{,}803 & 4{,}201 (23.8\%) & 1{,}312 (7.4\%) &   648 (3.7\%) &   642 (3.6\%) \\
client\,4 & 10{,}385 & 6{,}353 (23.5\%) & 2{,}150 (8.0\%) &   918 (3.4\%) &   964 (3.6\%) \\
client\,5 & 11{,}736 & 7{,}144 (23.4\%) & 2{,}384 (7.8\%) & 1{,}079 (3.5\%) & 1{,}129 (3.7\%) \\
\bottomrule
\end{tabular}%
}
\end{table}

\begin{table}[!htbp]
\centering
\caption{Distribution of organic OOD artifacts across \textbf{FedEMBED} clients. Values are reported as count (ratio \%); ratios are computed over each scanner's total image count. The heterogeneous distribution demonstrates the non-IID nature of clinical OOD.}
\label{tab:fedembed_ood}
\resizebox{\linewidth}{!}{%
\begin{tabular}{lc ccccc}
\toprule
\multirow{2}{*}{\textbf{Client (scanner)}} & \multirow{2}{*}{\textbf{Total OOD}} & \multicolumn{5}{c}{\textbf{Specific artifact breakdown}} \\
\cmidrule(lr){3-7}
 & & \textbf{Implants} & \textbf{Compression} & \textbf{Devices} & \textbf{Triangle} & \textbf{Circle} \\
\midrule
Selenia Dimensions & 33,804 (13.12\%) & 7,341 (2.85\%) & 239 (0.09\%) & 680 (0.26\%) & 3,705 (1.44\%) & 21,839 (8.48\%) \\
Senograph 2000D ADS & 1,421 (11.06\%) & 761 (5.92\%) & 71 (0.55\%) & 36 (0.28\%) & 212 (1.65\%) & 341 (2.65\%) \\
Lorad Selenia & 1,258 (12.37\%) & 424 (4.17\%) & 8 (0.08\%) & 29 (0.29\%) & 200 (1.97\%) & 597 (5.87\%) \\
Clearview CSm & 382 (4.83\%) & 1 (0.01\%) & 0 (0.00\%) & 7 (0.09\%) & 6 (0.08\%) & 368 (4.65\%) \\
\bottomrule
\end{tabular}%
}
\end{table}

\section{Implementation details}\label{ssec:openpath}

\textbf{Preprocessing.} Inputs follow the public protocol of each benchmark. FedISIC uses the entire dermoscopy image under the FLamby protocol~\citep{ogier2022flamby}: the shorter edge is resized to 224\,px with the aspect ratio preserved, followed by color-constancy normalization. FedCamelyon is patch-based: whole-slide images are tiled into $96\times96\times3$ patches following HarmoFL~\citep{Jiang_Wang_Dou_2022}. FedEMBED operates on the full breast following Mammo-CLIP~\citep{ghosh2024mammo}: the background is removed and the image resized to $1024\times768$. \texttt{FIDAL} itself is resolution-agnostic, since selection operates on model embeddings.

\textbf{Optimization.} All models train with AdamW (base learning rate 5e-4, batch size 32; learning rate 3e-4 for FedEMBED), except \texttt{EOAL}, which uses SGD (learning rate 1e-2, momentum 0.9) with a StepLR scheduler. Weight decay is 5e-4 (1e-4 for FedCamelyon). The evidential loss in \texttt{FEAL} uses a KL weight of 1e-2, as in \citet{chen2024think}; all other baselines use their original hyperparameters where applicable. \texttt{FIDAL} adds no optimizer-side hyperparameters beyond $(\lambda_{\text{div}},\lambda_{\text{OOD}})$: the gate threshold and the diversity threshold $\tau_{\text{ID}}$ are set adaptively as described in main-text Section~3, and no manual percentile cut-off is used.

\textbf{Federated OpenPath (\texttt{OpenPath*}).} The centralized \texttt{OpenPath}~\citep{zhong2025openpath} pipeline is run independently on every client: each client scores its own unlabeled pool with a frozen BiomedCLIP encoder (image embeddings precomputed and cached per dataset) against GPT-4-generated text prompts, following the original $M{=}8$ OOD-prompt protocol. The per-dataset prompt sets are:
\begin{itemize}
    \item \textbf{FedISIC}: 8 ID dermoscopy class prompts (``A dermoscopy image of \{Melanoma, Melanocytic nevus, Basal cell carcinoma, Actinic keratosis, Benign keratosis, Dermatofibroma, Vascular lesion, Squamous cell carcinoma\}'') and 8 OOD prompts for plausible non-target lesion types (Seborrheic keratosis, Lentigo, Dermatofibrosarcoma protuberans, Atypical melanocytic nevus, Spitz nevus, Blue nevus, Neurofibroma, Xanthoma).
    \item \textbf{FedCamelyon}: 2 ID histopathology prompts (normal lymph-node tissue, metastatic tumor tissue) and 8 OOD prompts for non-target tissue categories (lymphoid follicles, sinus histiocytosis, fibrous capsule / trabeculae, adipose tissue, blood vessels / lymphatic channels, necrosis / debris, inflammatory infiltrates, benign epithelial structures).
    \item \textbf{FedEMBED}: 4 ID BI-RADS density prompts (``A mammogram showing \{fatty, scattered fibroglandular, heterogeneously dense, extremely dense\} breast tissue'', BI-RADS A/B/C/D) and 8 OOD prompts for non-density acquisition categories (post-surgical changes, breast implant, biopsy markers / clips, post-radiation changes, technical-quality artifacts, non-breast anatomy / external artifacts, duplicate or prior-exam image, magnification / spot-compression view).
\end{itemize}

\section{Per-client Pareto trajectories}\label{ssec:perclient}

\begin{figure}[!htbp]
    \centering
    \includegraphics[width=\linewidth]{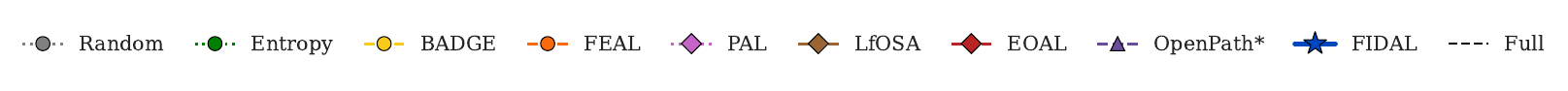}\\[0.4em]
    \textbf{FedISIC}\\[0.0em]
    \includegraphics[width=\linewidth]{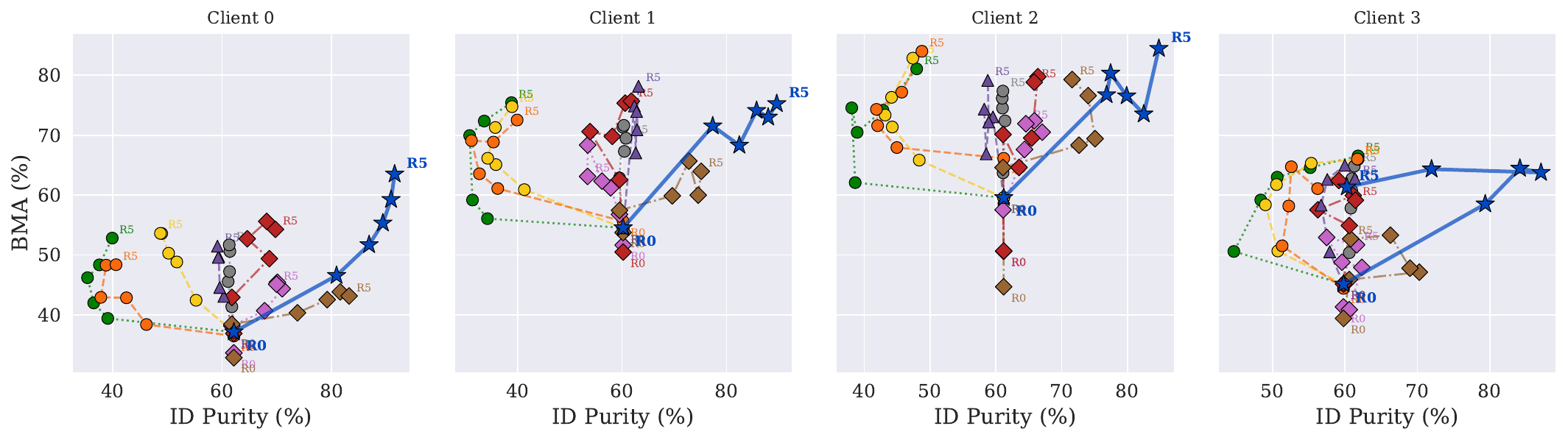}\\[0.3em]
    \textbf{FedCamelyon}\\[0.0em]
    \includegraphics[width=\linewidth]{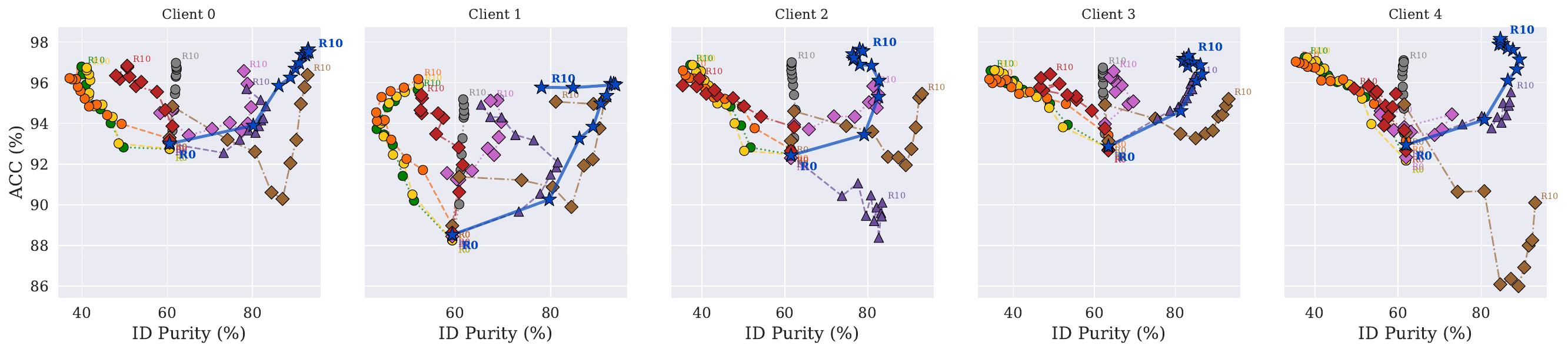}\\[0.3em]
    \textbf{FedEMBED}\\[0.0em]
    \includegraphics[width=\linewidth]{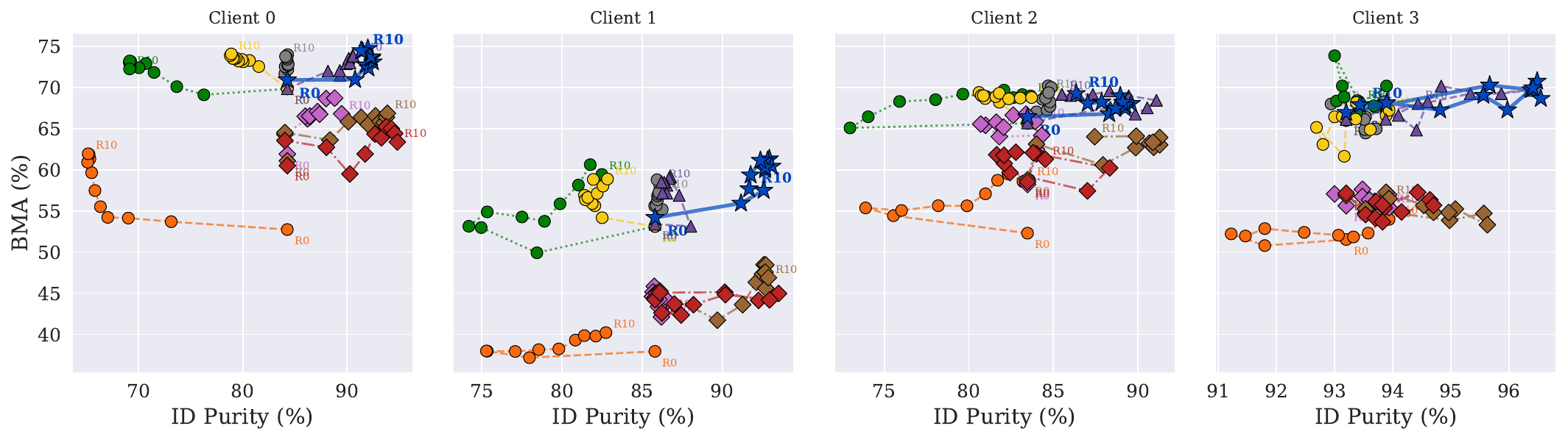}
    \caption{\textbf{Per-client Pareto: classification metric versus ID purity.} Each path is one method's (ID purity, BMA/ACC) trajectory across AL rounds $0\to R_{\max}$ ($R_{\max}{=}5$ for FedISIC, $10$ otherwise), decomposed by client (one panel per client, one block per dataset); upper-right is better. The aggregate gain of main-text Fig.~3 is not driven by a single center.}
    \label{fig:perclient_pareto_global}
\end{figure}

Fig.~\ref{fig:perclient_pareto_global} decomposes the federation-wide Pareto view of main-text Fig.~3 by client. On each of the four FedISIC clinics, \texttt{FIDAL} either tops the trajectory or matches the best baseline by the final round despite the marked label- and class-distribution heterogeneity across centers, so the aggregate gain is not an artifact of averaging over a favorable clinic.

\section{Per-round composition of the labeled pool}\label{ssec:composition_rounds}

\begin{figure}[!htbp]
    \centering
    \includegraphics[width=\linewidth]{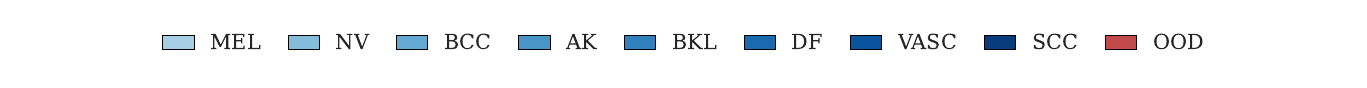}\\[-0.2em]
    \includegraphics[width=\linewidth]{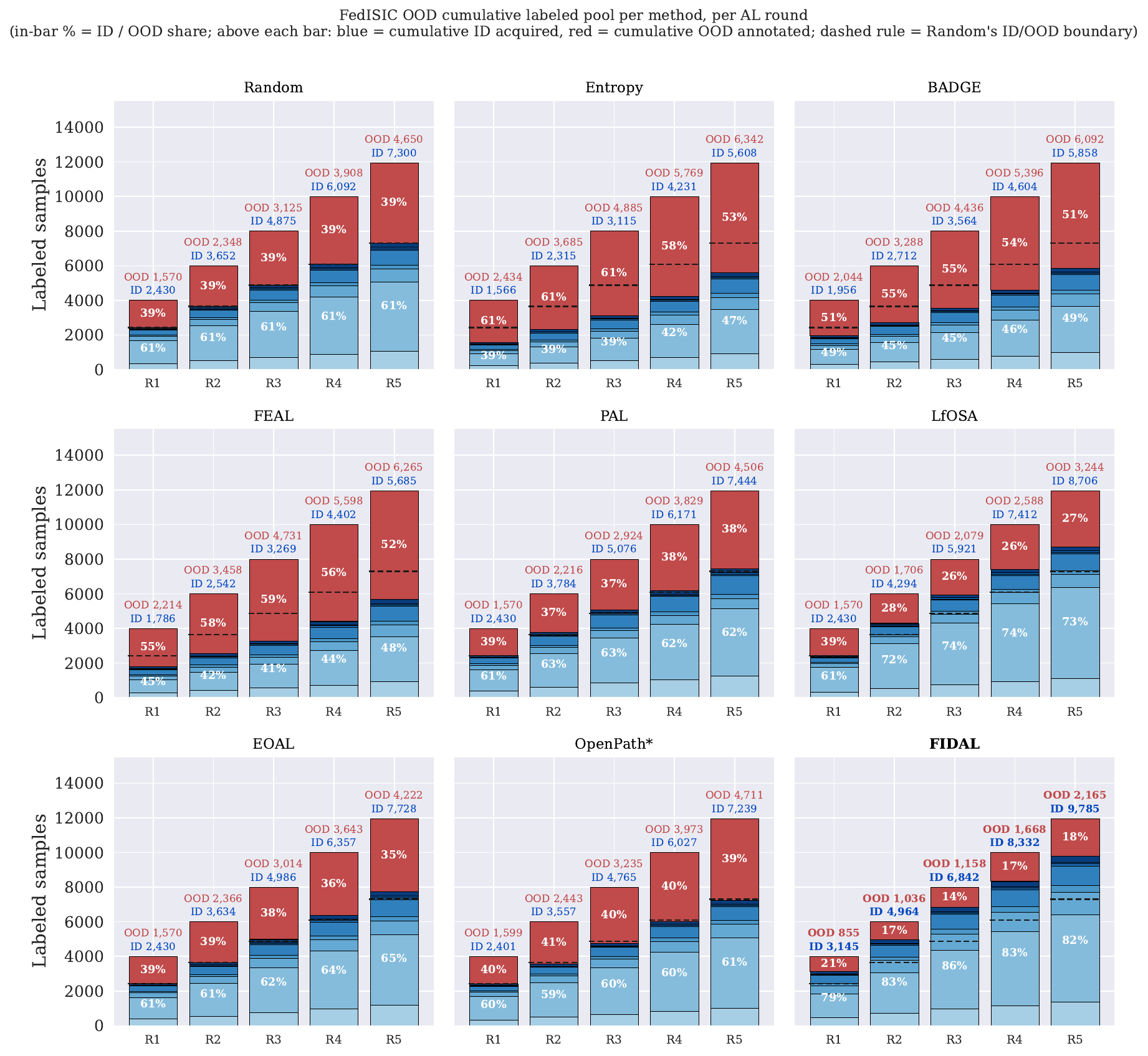}
    \caption{\textbf{FedISIC OOD} ($B{=}500$, $E{=}100$, 3 seeds). Per-method cumulative labeled-pool composition across AL rounds $R{=}1\dots5$ (the final acquisition is logged under $R{=}4$ and shown here as $R{=}5$, matching main-text Table~1): stacked ID (blue shades) and OOD (red), summed across the 4 clients. The number above each bar is the total count; in-bar percentages give the ID / OOD share. Above each bar, blue \textbf{ID\,N} and red \textbf{OOD\,N} give the cumulative number of ID and OOD samples that method has sent for annotation up to that round; their sum is the cumulative labeled pool, which is identical across methods except where a client exhausts its local pool (\texttt{EOAL} on FedEMBED). The dashed rule marks the canonical AL control \texttt{Random}: a red segment beginning \emph{above} the rule means the method acquired more ID, and wasted less budget on OOD, than Random at that round. Higher blue and lower red at the same total height is the desired behavior.}
    \label{fig:isic-perround-composition}
\end{figure}

\begin{figure}[!htbp]
    \centering
    \includegraphics[width=\linewidth]{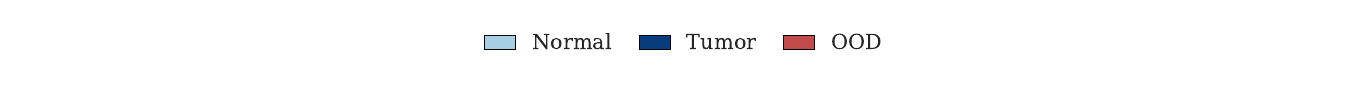}\\[-0.2em]
    \includegraphics[width=\linewidth]{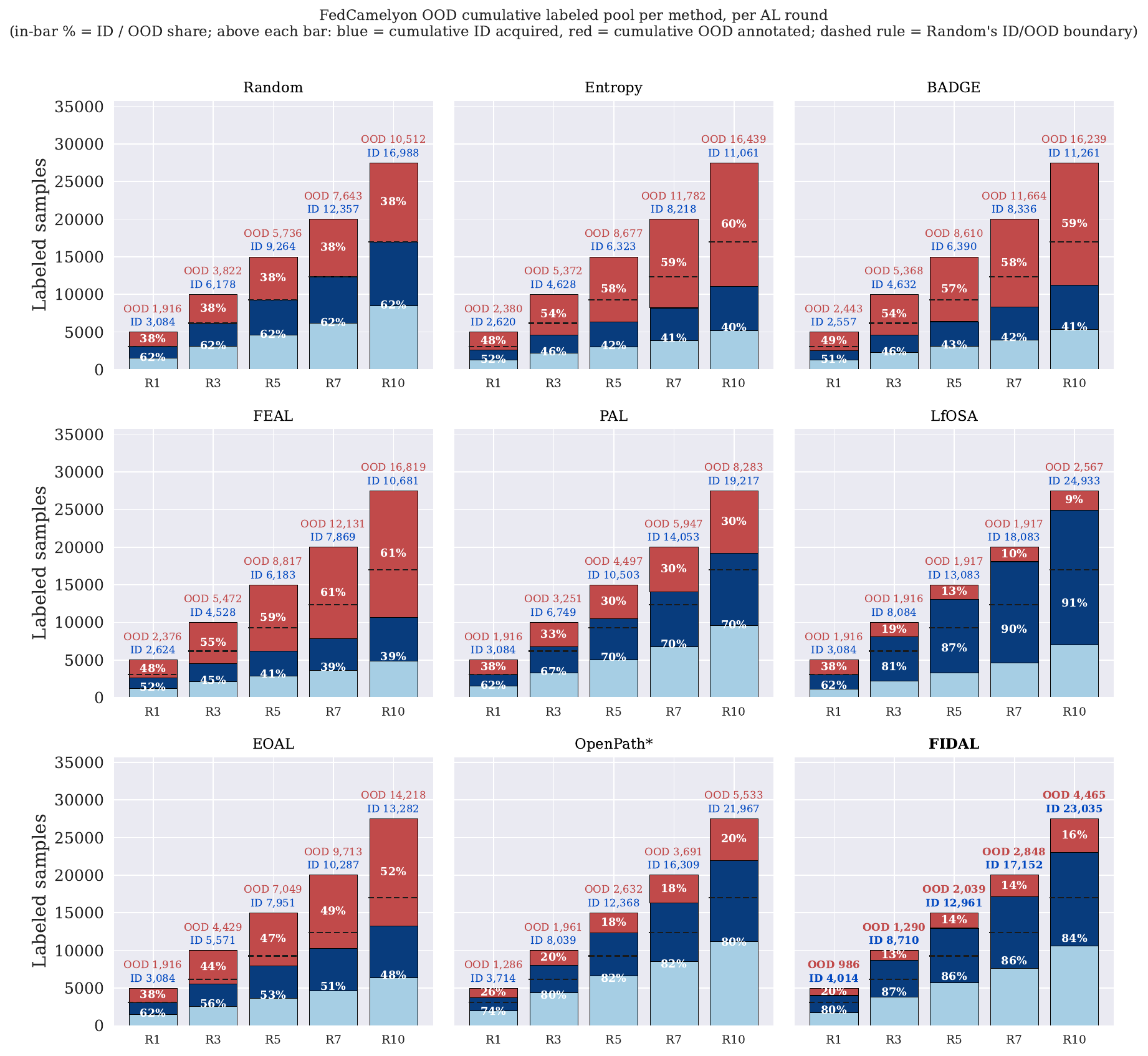}
    \caption{\textbf{FedCamelyon OOD} ($B{=}500$, $E{=}100$, mean across 3 seeds). Per-method composition of the cumulative labeled pool across AL rounds $R\in\{1,3,5,7,10\}$; layout and annotations as in Fig.~\ref{fig:isic-perround-composition}. Only two ID classes (\textbf{Normal}, \textbf{Tumor}) plus OOD.}
    \label{fig:camelyon-perround-composition}
\end{figure}

\begin{figure}[!htbp]
    \centering
    \includegraphics[width=\linewidth]{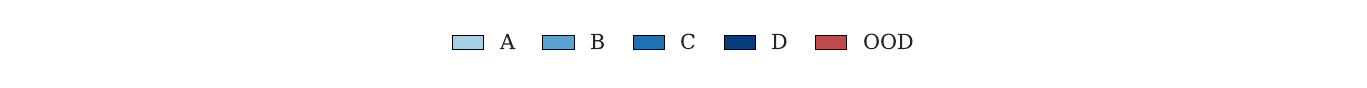}\\[-0.2em]
    \includegraphics[width=\linewidth]{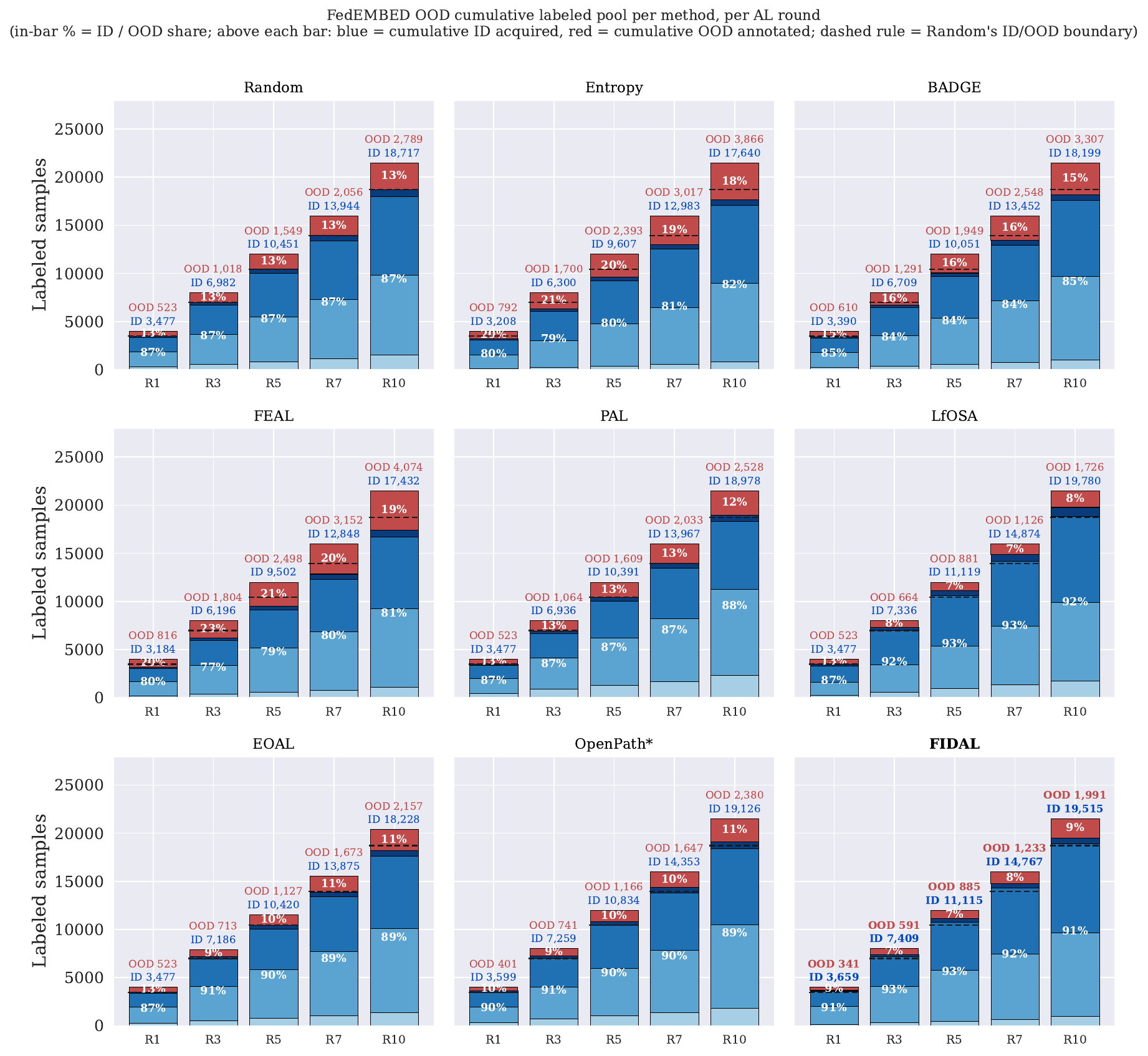}
    \caption{\textbf{FedEMBED OOD} ($B{=}500$, $E{=}50$, mean across 3 seeds). Per-method composition of the cumulative labeled pool across AL rounds $R\in\{1,3,5,7,10\}$; layout and annotations as in Fig.~\ref{fig:isic-perround-composition}. Four BI-RADS density classes (\textbf{A}, \textbf{B}, \textbf{C}, \textbf{D}) plus OOD. Because organic OOD prevalence is low ($4.8$--$13.1\%$ across clients), the red segments are thin and the ID\,/\,OOD differences between methods are correspondingly smaller than on the controlled benchmarks.}
    \label{fig:embed-perround-composition}
\end{figure}

Figs.~\ref{fig:isic-perround-composition}--\ref{fig:embed-perround-composition} show, for each method, how the cumulative labeled pool is built up round by round (ID classes in blue shades, OOD in red); main-text Section~4.3 gives the headline OOD shares. Two details are visible only in the trajectories. On FedISIC, \texttt{FIDAL}'s cumulative OOD share peaks at $21.4\%$ after the first query round and then settles to $14$--$18\%$. On FedCamelyon, from $R{\ge}5$ \texttt{LfOSA}'s aggressive GMM filter matches or exceeds \texttt{FIDAL} on raw ID count, yet \texttt{FIDAL} still avoids more than $11{,}000$ OOD queries at $R{=}10$ relative to a low-purity baseline such as \texttt{Entropy}.

\section{Budget and local-epoch variation}\label{ssec:budget_epoch}

Tables~\ref{tab:fedisic_combined}--\ref{tab:fedembed_combined_global} report the full stress test under varying per-round budget $B$ (at the headline $E$) and local epochs $E$ (at $B{=}500$) for the three benchmarks at their native OOD configuration; main-text Section~4.3 summarizes them. Conventions: FedISIC and FedCamelyon report each method's best AL round (peak round in parentheses for FedISIC), FedEMBED the final round $R{=}10$; the second line of each cell is the cumulative ID purity at that round; \textbf{Avg.} averages the seven distinct $(B,E)$ cells, since the $B{=}500$ cell at the headline $E$ is shared by both sweeps. Relative to its own peak, \texttt{FIDAL} retains $86\%$ of BMA at $E{=}2$ on FedISIC, $92\%$ of ACC on FedCamelyon, and $98\%$ of BMA on FedEMBED.

\begin{table}[!htbp]
\centering
\caption{\textbf{FedISIC} OOD budget and local-epoch variation: best-AL-round BMA (\%, top of each cell with peak round in parentheses) and cumulative ID purity (\%, bottom). Mean$\pm$std across 3 seeds. \texttt{FIDAL} at $(\lambda_{\text{div}},\lambda_{\text{OOD}}){=}(0.5,0.5)$. Best BMA / ID purity per column in \textbf{bold} / \underline{underlined}. \textbf{Avg.} averages over the seven distinct configurations.}
\label{tab:fedisic_combined}
\resizebox{\textwidth}{!}{%
\begin{tabular}{l @{\hspace{10pt}} cccc @{\hspace{10pt}} cccc @{\hspace{10pt}} c}
\toprule
\multirow{2}{*}{\textbf{Method}} & \multicolumn{4}{c}{\textbf{Budget variation} ($E{=}100$)} & \multicolumn{4}{c}{\textbf{Local-epoch variation} ($B{=}500$)} & \multirow{2}{*}{\textbf{Avg.}} \\
\cmidrule(lr){2-5}\cmidrule(lr){6-9}
 & $B{=}50$ & $B{=}150$ & $B{=}300$ & $B{=}500$ & $E{=}2$ & $E{=}10$ & $E{=}50$ & $E{=}100$ & \\
\midrule
Random & 42.28\,$\pm$3.21~(R5) & 54.74\,$\pm$2.50~(R5) & 60.97\,$\pm$3.09~(R5) & 65.83\,$\pm$2.66~(R5) & 58.25\,$\pm$2.89~(R4) & 65.72\,$\pm$1.77~(R5) & 65.15\,$\pm$1.41~(R4) & 65.83\,$\pm$2.66~(R5) & 58.99 \\
  & 61.9 & 60.6 & 61.3 & 61.1 & 60.9 & 61.1 & 60.9 & 61.1 & 61.1 \\
Entropy & 30.47\,$\pm$0.72~(R3) & 48.10\,$\pm$1.58~(R5) & 60.44\,$\pm$2.56~(R5) & 68.96\,$\pm$1.11~(R5) & 59.03\,$\pm$1.27~(R5) & 64.11\,$\pm$1.35~(R5) & 66.73\,$\pm$0.98~(R5) & 68.96\,$\pm$1.11~(R5) & 56.83 \\
  & 17.0 & 20.3 & 35.8 & 46.9 & 44.3 & 45.0 & 46.2 & 46.9 & 36.5 \\
BADGE & 45.87\,$\pm$4.38~(R5) & 56.45\,$\pm$4.40~(R5) & 64.58\,$\pm$2.90~(R5) & 69.29\,$\pm$2.23~(R5) & 59.94\,$\pm$1.64~(R4) & 65.10\,$\pm$1.90~(R5) & 67.02\,$\pm$1.51~(R5) & 69.29\,$\pm$2.23~(R5) & 61.18 \\
  & 48.9 & 46.1 & 47.5 & 49.0 & 56.0 & 51.5 & 49.7 & 49.0 & 49.8 \\
FEAL & 30.16\,$\pm$0.89~(R1) & 49.39\,$\pm$2.35~(R4) & 61.05\,$\pm$3.48~(R5) & 67.73\,$\pm$0.75~(R5) & 32.83\,$\pm$2.20~(R5) & 55.80\,$\pm$1.38~(R5) & 58.46\,$\pm$0.97~(R5) & 67.73\,$\pm$0.75~(R5) & 50.77 \\
  & 33.5 & 26.3 & 38.9 & 47.6 & 42.4 & 44.4 & 47.3 & 47.6 & 40.1 \\
\midrule
PAL & 40.94\,$\pm$6.79~(R5) & 51.93\,$\pm$1.44~(R5) & 53.79\,$\pm$3.59~(R4) & 59.77\,$\pm$0.18~(R4) & 36.27\,$\pm$3.28~(R5) & 58.52\,$\pm$1.79~(R5) & 52.82\,$\pm$1.00~(R4) & 59.77\,$\pm$0.18~(R4) & 50.58 \\
  & 68.6 & 63.7 & 64.3 & 61.7 & 67.2 & 63.2 & 60.8 & 61.7 & 64.2 \\
LfOSA & 35.56\,$\pm$2.82~(R5) & 44.95\,$\pm$2.33~(R5) & 54.03\,$\pm$0.81~(R5) & 59.77\,$\pm$1.37~(R5) & 38.55\,$\pm$1.84~(R3) & 53.23\,$\pm$2.66~(R5) & 54.02\,$\pm$4.39~(R4) & 59.77\,$\pm$1.37~(R5) & 48.59 \\
  & 78.4 & 80.0 & 77.9 & 72.9 & 77.0 & 74.3 & 75.2 & 72.9 & 76.5 \\
EOAL & 39.42\,$\pm$0.34~(R4) & 55.74\,$\pm$2.82~(R4) & 65.65\,$\pm$1.29~(R4) & 68.28\,$\pm$0.73~(R4) & 37.87\,$\pm$2.07~(R5) & 58.68\,$\pm$1.39~(R5) & 59.06\,$\pm$4.04~(R4) & 68.28\,$\pm$0.73~(R4) & 54.96 \\
  & 48.5 & 67.4 & 59.0 & 63.6 & 58.4 & 54.0 & 58.0 & 63.6 & 58.4 \\
OpenPath* & 37.65\,$\pm$1.97~(R5) & 51.33\,$\pm$2.84~(R5) & 63.20\,$\pm$0.62~(R5) & 67.89\,$\pm$0.85~(R5) & 59.58\,$\pm$2.59~(R5) & 65.09\,$\pm$1.87~(R5) & 63.41\,$\pm$1.24~(R5) & 67.89\,$\pm$0.85~(R5) & 58.31 \\
  & 59.4 & 59.7 & 59.7 & 60.6 & 60.6 & 60.6 & 60.6 & 60.6 & 60.2 \\
\midrule
\textbf{FIDAL} & \textbf{50.71\,$\pm$1.62~(R5)} & \textbf{64.94\,$\pm$1.49~(R5)} & \textbf{69.08\,$\pm$2.66~(R5)} & \textbf{71.14\,$\pm$0.54~(R5)} & \textbf{61.47\,$\pm$0.54~(R4)} & \textbf{68.49\,$\pm$0.88~(R4)} & \textbf{69.54\,$\pm$0.85~(R3)} & \textbf{71.14\,$\pm$0.54~(R5)} & \textbf{65.05} \\
  & \underline{93.2} & \underline{88.7} & \underline{88.1} & \underline{81.9} & \underline{83.0} & \underline{83.2} & \underline{85.0} & \underline{81.9} & \underline{86.2} \\
\bottomrule
\end{tabular}}
\end{table}

\begin{table}[!htbp]
\centering
\caption{\textbf{FedCamelyon} OOD budget and local-epoch variation: best-AL-round ACC (\%, top of each cell) and cumulative ID purity (\%, bottom). Mean$\pm$std across 3 seeds. \texttt{FIDAL} at $(\lambda_{\text{div}},\lambda_{\text{OOD}}){=}(1.0,1.0)$. Best ACC / ID purity per column in \textbf{bold} / \underline{underlined}.}
\label{tab:fedcamelyon_combined_global}
\resizebox{\textwidth}{!}{%
\begin{tabular}{l @{\hspace{10pt}} cccc @{\hspace{10pt}} cccc @{\hspace{10pt}} c}
\toprule
\multirow{2}{*}{\textbf{Method}} & \multicolumn{4}{c}{\textbf{Budget variation} ($E{=}100$)} & \multicolumn{4}{c}{\textbf{Local-epoch variation} ($B{=}500$)} & \multirow{2}{*}{\textbf{Avg.}} \\
\cmidrule(lr){2-5}\cmidrule(lr){6-9}
 & $B{=}50$ & $B{=}150$ & $B{=}300$ & $B{=}500$ & $E{=}2$ & $E{=}10$ & $E{=}50$ & $E{=}100$ & \\
\midrule
Random & 91.55\,$\pm$0.28 & 94.40\,$\pm$0.32 & 95.82\,$\pm$0.05 & 96.53\,$\pm$0.13 & 88.80\,$\pm$0.32 & 92.50\,$\pm$0.38 & 95.41\,$\pm$0.12 & 96.53\,$\pm$0.13 & 93.57 \\
  & 62.3 & 59.9 & 60.9 & 61.8 & 61.8 & 61.8 & 61.8 & 61.8 & 61.5 \\
Entropy & 91.49\,$\pm$0.26 & 93.85\,$\pm$0.03 & 95.38\,$\pm$0.07 & 96.63\,$\pm$0.04 & 89.20\,$\pm$0.29 & 92.49\,$\pm$0.19 & 95.42\,$\pm$0.31 & 96.63\,$\pm$0.04 & 93.49 \\
  & 45.1 & 43.2 & 41.1 & 40.2 & 49.2 & 49.2 & 43.1 & 40.2 & 44.4 \\
BADGE & 91.43\,$\pm$0.70 & 94.08\,$\pm$0.20 & 95.50\,$\pm$0.12 & 96.60\,$\pm$0.09 & 88.68\,$\pm$0.26 & 92.32\,$\pm$0.38 & 95.53\,$\pm$0.17 & 96.60\,$\pm$0.09 & 93.45 \\
  & 46.9 & 43.8 & 41.9 & 40.9 & 49.9 & 50.2 & 44.5 & 40.9 & 45.4 \\
FEAL & \textbf{92.20\,$\pm$0.16} & 94.36\,$\pm$0.04 & 95.59\,$\pm$0.14 & 96.49\,$\pm$0.09 & 88.92\,$\pm$0.87 & 92.58\,$\pm$0.09 & 95.35\,$\pm$0.08 & 96.49\,$\pm$0.09 & 93.64 \\
  & 51.2 & 42.9 & 40.5 & 38.8 & 48.9 & 49.1 & 43.3 & 38.8 & 45.0 \\
\midrule
PAL & 88.54\,$\pm$0.41 & 92.72\,$\pm$1.60 & 93.26\,$\pm$0.70 & 95.67\,$\pm$0.39 & 87.36\,$\pm$0.14 & 90.92\,$\pm$0.64 & 93.58\,$\pm$0.92 & 95.67\,$\pm$0.39 & 91.72 \\
  & 58.1 & 62.6 & 67.0 & 69.9 & 59.9 & 61.6 & 67.2 & 69.9 & 63.8 \\
LfOSA & 87.33\,$\pm$0.06 & 90.89\,$\pm$0.27 & 93.12\,$\pm$0.60 & 94.41\,$\pm$0.70 & 87.22\,$\pm$0.57 & 90.87\,$\pm$0.33 & 93.31\,$\pm$0.39 & 94.41\,$\pm$0.70 & 91.02 \\
  & 62.2 & 60.5 & \underline{92.9} & \underline{90.7} & 61.7 & 87.2 & 61.7 & \underline{90.7} & 73.8 \\
EOAL & 91.14\,$\pm$0.20 & 94.20\,$\pm$0.43 & 95.06\,$\pm$0.49 & 96.15\,$\pm$0.24 & 87.89\,$\pm$0.76 & 92.47\,$\pm$0.11 & 95.50\,$\pm$0.13 & 96.15\,$\pm$0.24 & 93.20 \\
  & 50.3 & 54.4 & 54.8 & 48.3 & 55.2 & 61.7 & 55.6 & 48.3 & 54.3 \\
OpenPath* & 85.28\,$\pm$1.53 & 89.16\,$\pm$0.48 & 90.94\,$\pm$0.43 & 94.61\,$\pm$0.41 & 86.76\,$\pm$0.74 & 89.97\,$\pm$0.33 & 91.65\,$\pm$0.45 & 94.61\,$\pm$0.41 & 89.77 \\
  & 62.0 & 61.5 & 61.3 & 79.9 & 61.4 & 61.4 & 61.4 & 79.9 & 64.1 \\
\midrule
\textbf{FIDAL} & 91.82\,$\pm$0.45 & \textbf{95.52\,$\pm$0.21} & \textbf{96.95\,$\pm$0.01} & \textbf{97.34\,$\pm$0.12} & \textbf{89.77\,$\pm$0.47} & \textbf{92.97\,$\pm$0.15} & \textbf{96.10\,$\pm$0.06} & \textbf{97.34\,$\pm$0.12} & \textbf{94.35} \\
  & \underline{96.1} & \underline{95.5} & 87.9 & 83.8 & \underline{88.6} & \underline{89.5} & \underline{87.4} & 83.8 & \underline{89.8} \\
\bottomrule
\end{tabular}}
\end{table}

\begin{table}[!htbp]
\centering
\caption{\textbf{FedEMBED} budget and local-epoch variation: final-AL-round ($R{=}10$) BMA (\%, top) and cumulative ID purity (\%, bottom). Mean$\pm$std across 3 seeds. \texttt{FIDAL} at $(\lambda_{\text{div}},\lambda_{\text{OOD}}){=}(0.5,5.0)$. Best BMA / ID purity per column in \textbf{bold} / \underline{underlined} at the final round $R{=}10$.}
\label{tab:fedembed_combined_global}
\resizebox{\textwidth}{!}{%
\begin{tabular}{l @{\hspace{10pt}} cccc @{\hspace{10pt}} cccc @{\hspace{10pt}} c}
\toprule
\multirow{2}{*}{\textbf{Method}} & \multicolumn{4}{c}{\textbf{Budget variation} ($E{=}50$)} & \multicolumn{4}{c}{\textbf{Local-epoch variation} ($B{=}500$)} & \multirow{2}{*}{\textbf{Avg.}} \\
\cmidrule(lr){2-5}\cmidrule(lr){6-9}
 & $B{=}50$ & $B{=}150$ & $B{=}300$ & $B{=}500$ & $E{=}2$ & $E{=}10$ & $E{=}25$ & $E{=}50$ & \\
\midrule
Random & 61.55\,$\pm$1.01 & 65.47\,$\pm$1.00 & 66.56\,$\pm$0.22 & 67.19\,$\pm$0.57 & 64.82\,$\pm$1.65 & 66.73\,$\pm$1.01 & 67.77\,$\pm$1.30 & 67.19\,$\pm$0.57 & 65.73 \\
  & 86.9 & 87.5 & 87.3 & 87.0 & 87.0 & 87.0 & 87.0 & 87.0 & 87.1 \\
Entropy & 59.27\,$\pm$2.01 & 64.28\,$\pm$0.24 & 66.96\,$\pm$0.28 & 68.19\,$\pm$0.40 & 65.95\,$\pm$0.48 & 68.19\,$\pm$1.45 & \textbf{68.07\,$\pm$0.81} & 68.19\,$\pm$0.40 & 65.84\\
  & 77.7 & 74.3 & 79.4 & 82.0 & 84.9 & 82.8 & 82.0 & 82.0 & 80.4\\
BADGE & 64.02\,$\pm$1.44 & 64.44\,$\pm$1.28 & 66.41\,$\pm$0.74 & 68.43\,$\pm$0.64 & 65.82\,$\pm$0.50 & 67.50\,$\pm$0.54 & 67.69\,$\pm$1.50 & 68.43\,$\pm$0.64 & 66.33 \\
  & 84.7 & 82.8 & 84.0 & 84.6 & 85.5 & 84.9 & 84.9 & 84.6 & 84.5 \\
FEAL & 42.12\,$\pm$0.67 & 47.09\,$\pm$0.88 & 50.55\,$\pm$0.21 & 53.72\,$\pm$0.21 & 48.13\,$\pm$0.30 & 48.95\,$\pm$0.62 & 51.68\,$\pm$0.21 & 53.72\,$\pm$0.21 & 48.89 \\
  & 75.9 & 72.7 & 77.6 & 81.1 & 82.7 & 81.5 & 81.1 & 81.1 & 78.9 \\
\midrule
PAL & 55.46\,$\pm$1.01 & 56.69\,$\pm$1.95 & 58.14\,$\pm$0.89 & 58.74\,$\pm$0.60 & 54.12\,$\pm$1.03 & 58.37\,$\pm$0.86 & 58.54\,$\pm$1.13 & 58.74\,$\pm$0.60 & 57.15 \\
  & 79.3 & 83.4 & 86.7 & 88.2 & 83.1 & 87.3 & 88.0 & 88.2 & 85.1 \\
LfOSA & 54.46\,$\pm$0.62 & 57.96\,$\pm$1.92 & 59.56\,$\pm$0.37 & 58.80\,$\pm$0.63 & 55.01\,$\pm$0.22 & 56.14\,$\pm$1.11 & 58.15\,$\pm$0.78 & 58.80\,$\pm$0.63 & 57.15 \\
  & 94.0 & \underline{94.9} & \underline{93.8} & \underline{92.0} & 90.7 & \underline{92.5} & \underline{92.5} & \underline{92.0} & \underline{92.9} \\
EOAL & 55.04\,$\pm$0.83 & 57.30\,$\pm$0.78 & 57.70\,$\pm$0.75 & 57.27\,$\pm$0.31 & 55.25\,$\pm$0.93 & 57.45\,$\pm$0.50 & 58.03\,$\pm$0.21 & 57.27\,$\pm$0.31 & 56.86 \\
  & 91.2 & 94.1 & 90.4 & 89.4 & 81.2 & 83.2 & 87.1 & 89.4 & 88.1 \\
OpenPath* & \textbf{64.92\,$\pm$1.24} & 67.57\,$\pm$1.15 & 67.17\,$\pm$0.52 & 67.55\,$\pm$0.64 & 66.29\,$\pm$0.43 & 66.92\,$\pm$0.67 & 66.69\,$\pm$0.05 & 67.55\,$\pm$0.64 & 66.73 \\
  & 91.2 & 91.3 & 89.8 & 88.9 & 88.9 & 88.9 & 88.9 & 88.9 & 89.7 \\
\midrule
\textbf{FIDAL} & 60.90\,$\pm$1.59 & \textbf{67.77\,$\pm$0.72} & \textbf{68.13\,$\pm$0.85} & \textbf{68.21\,$\pm$0.31} & \textbf{66.82\,$\pm$0.31} & \textbf{68.31\,$\pm$0.33} & 68.03\,$\pm$0.56 & \textbf{68.21\,$\pm$0.31} & \textbf{66.88} \\
  & \underline{95.5} & 94.7 & 93.4 & 90.7 & \underline{91.1} & 91.3 & 91.0 & 90.7 & 92.5 \\
\bottomrule
\end{tabular}}
\end{table}

\section{Extended analysis of per-class recall on FedISIC (main-text Fig.~6)}\label{ssec:rare}

Main-text Fig.~6 reports per-class recall on FedISIC and main-text Section~5.1 gives the per-class gaps. This section documents the protocol and its caveats. FedISIC was chosen for the class-wise analysis because it shows the largest between-method accuracy gap and remains furthest from saturation. Recall is computed per class at a common AL round ($R{=}3$) for budgets $B\in\{50,150,300,500\}$ at $E{=}100$ and averaged over 3 seeds; the annotated gap is \texttt{FIDAL} minus the best non-FIDAL baseline for that class and budget. Two caveats apply. First, the rare-first pattern is consistent with under-selection of the rare classes by the baselines, whose batches contain up to $61\%$ OOD on FedISIC (Fig.~\ref{fig:isic-perround-composition}), but it does not establish that rare-class images were rejected specifically as artifacts. Second, the per-class leads narrow as the budget grows because the baselines recover the scarce classes on their own, so the largest advantages, including the largest gaps over the open-set baselines, occur at lower budgets, where each annotation has greater influence.

\FloatBarrier